\documentclass[10pt,twocolumn,letterpaper]{article}

\newcommand{\dv}[0]{\mathbf{d}}
\newcommand{\fv}[0]{\mathbf{f}}

\newcommand{\Imat}[0]{\mathbf{I}}
\newcommand{\kv}[0]{\mathbf{k}}

\newcommand{\mv}[0]{\mathbf{m}}
\newcommand{\ov}[0]{\mathbf{o}}
\newcommand{\pv}[0]{\mathbf{p}}
\newcommand{\qv}[0]{\mathbf{q}}
\newcommand{\rv}[0]{\mathbf{r}}

\newcommand{\vv}[0]{\mathbf{v}}
\newcommand{\Wm}[0]{\mathbf{W}}

\newcommand{\zv}[0]{\mathbf{z}}

\newcommand{\fcnn}[0]{F_{{C}}}
\newcommand{\fenc}[0]{F_{{E}}}
\newcommand{\fdec}[0]{F_{{D}}}

\newcommand{\biasweight}[0]{\gamma}

\newcommand{\plucker}[0]{Plücker }  % keep the space at the right end
\newcommand{\red}[1]{{\color{red}#1}}

\usepackage[pagenumbers]{cvpr}

\makeatletter
\@namedef{ver@everyshi.sty}{}
\makeatother
\usepackage{tikz}

\usepackage{graphicx}
\usepackage{amsmath}
\usepackage{amssymb}
\usepackage{booktabs}
\usepackage{multirow}
\usepackage{nicematrix}
\usepackage[title]{appendix}

\usepackage{lib}

\usepackage[pagebackref,breaklinks,colorlinks]{hyperref}

\usepackage[capitalize]{cleveref}
\crefname{section}{Sec.}{Secs.}
\Crefname{section}{Section}{Sections}
\Crefname{table}{Table}{Tables}
\crefname{table}{Tab.}{Tabs.}

\begin{document}

\title{Geometry-biased Transformers for Novel View Synthesis}

\vspace{-2mm}
\author{
Naveen Venkat\thanks{Equal Contribution}$\hspace{4pt}^{1}$ \quad Mayank Agarwal\footnotemark[1]$\hspace{4pt}^{1}$ \quad Maneesh Singh \quad Shubham Tulsiani$^1$\\
$^1$Carnegie Mellon University\\
{\tt\small \{nvenkat, mayankag, shubhtuls\}@cmu.edu, dr.maneesh.singh@ieee.org}
\\ {\tt \small \href{https://mayankgrwl97.github.io/gbt}{https://mayankgrwl97.github.io/gbt}}
}

\twocolumn[{
\maketitle
\renewcommand\twocolumn[1][]{#1}
\centering 
\vspace{-7mm}
\includegraphics[width=1.0\linewidth]{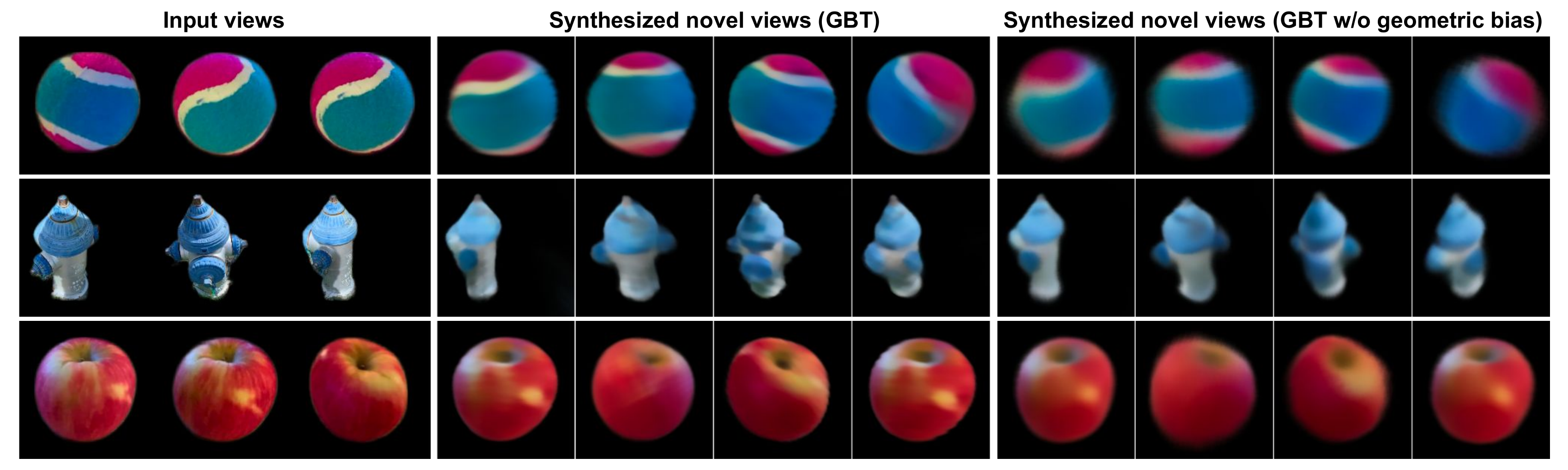}
\vspace{-8mm}
\captionof{figure}{
\small
   Given a small set of context images with known camera viewpoints (left), our Geometry-biased transformer (GBT) synthesizes novel views from arbitrary query viewpoints (middle). The use of global context ensures meaningful prediction despite large viewpoint variation, while the geometric bias allows more accurate inference compared to a baseline without such bias (right).
} 
\figlabel{teaser}
\vspace{3mm}
}]

\vspace{-2mm}
% \let\thefootnote\relax\footnotetext{* indicates equal contribution\\
% % \indent\indent Project Page: \texttt{\href{https://mayankgrwl97.github.io/gbt}{mayankgrwl97.github.io/gbt}}
% }

%%%%%%%%% ABSTRACT
\begin{abstract}
\vspace{-4mm}
We tackle the task of synthesizing novel views of an object given a few input images and associated camera viewpoints. Our work is inspired by recent `geometry-free' approaches where multi-view images are encoded as a (global) set-latent representation, which is then used to predict the color for arbitrary query rays. While this representation yields (coarsely) accurate images corresponding to novel viewpoints, the lack of geometric reasoning limits the quality of these outputs. To overcome this limitation, we propose `Geometry-biased Transformers' (GBTs) that incorporate geometric inductive biases in the set-latent representation-based inference to encourage multi-view geometric consistency. We induce the geometric bias by augmenting the dot-product attention mechanism to also incorporate 3D distances between rays associated with tokens as a learnable bias. We find that this, along with camera-aware embeddings as input, allows our models to generate significantly more accurate outputs. We validate our approach on the real-world CO3D dataset, where we train our system over 10 categories and evaluate its view-synthesis ability for novel objects as well as unseen categories. We empirically validate the benefits of the proposed geometric biases and show that our approach significantly improves over prior works. %The project page is available at: \texttt{\href{https://mayankgrwl97.github.io/gbt}{https://mayankgrwl97.github.io/gbt}}.
\end{abstract}

%%%%%%%%%%%%%%%%%%%%%%%%%%%

\section{Introduction}
\vspace{-2mm}
\label{introduction}
Given just a few images depicting an object, we humans can easily imagine its appearance from novel viewpoints. For instance, consider the first image of the hydrant shown in \figref{teaser} and imagine rotating it slightly anti-clockwise -- we intuitively understand that this would move the small outlet towards the front and right. We can also imagine rotating the hydrant further and know that the (currently occluded) central outlet will eventually become visible on the left. These examples serve to highlight that this task of novel-view synthesis requires both reasoning about geometric transformations \eg motion of the visible surfaces, as well as an understanding of the global structure \eg occlusions and symmetries to allow for realistic extrapolations. In this work, we develop an approach that incorporates both these to synthesize accurate novel views given only a sparse set of images of a previously unseen object.

\let\thefootnote\relax\footnotetext{* indicates equal contribution}

Recent advances in Neural Radiance Fields (NeRFs) \cite{mildenhall2020nerf} have led to numerous approaches that use these representations (and their variants) for obtaining remarkably detailed novel-view renderings. However, such methods typically optimize instance-specific representations using densely sampled multi-view observations, and cannot be directly leveraged for 3D inference from sparse input views. To enable generalizable inference from a few views, recent methods seek to instead predict radiance fields using the image projections of a query 3D point as conditioning. While using such geometric reprojection constraints allows accurate predictions in the close vicinity of observed views, this purely local conditioning mechanism fails to capture any global context \eg symmetries or correlated patterns. As a result, these approaches struggle to render views containing unobserved aspects or large viewpoint variations.

Our work is motivated by an alternate approach to generalizable view synthesis, where a geometry-free (global) scene representation is used to predict images from query viewpoints. Specifically, these methods form a set-latent representation from multiple input views and directly infer the color for a pixel for a query view (or equivalently a query ray) using attention-based mechanisms in the scene encoding and ray decoding process. Not only is this direct view synthesis more computationally efficient than volume rendering, but the set-latent representation also allows capturing global context as each ray can attend to all aspects of other views instead of just the projections of points along it. However, this `geometry-free' design comes at the cost of precision -- these methods cannot easily capture the details in input views, and while they can robustly capture the coarse structure, do not output high-quality renderings. 

In this work, we develop mechanisms to inject geometric biases in these set-latent representation-based approaches. Specifically, we propose Geometry-biased Transformers (GBTs) which consist of a ray-distance-based bias in the attention mechanism in Transformer layers. We show that these help guide the scene encoding and ray decoding stages to pay attention to relevant context, thereby enabling more accurate view synthesis. We benchmark our approach using the Co3D dataset ~\cite{reizenstein2021common} that comprises of challenging real-world captures across diverse categories. We show that our approach outperforms both,  projection-based radiance field prediction and set-latent representation-based view synthesis approaches, and also demonstrate our method's ability to generalize to unseen object categories.

%%%%%%%%%%%%%%%%%%%%%%%%%%%

\vspace{-2mm}
\section{Related Work}
\vspace{-2mm}
\label{related}
\paragraph{Instance-specific 3D Representations.}
Driven by the recent emergence of neural fields~\cite{mildenhall2020nerf}, a growing number of methods seek to accurately capture the details of a specific object or scene given multiple images. Leveraging either  volumetric~\cite{mildenhall2020nerf,kangle2021dsnerf,mueller2022instant,boss2022samurai,Niemeyer2021Regnerf,Jain_2021_ICCV,barron2022mipnerf360}, implicit~\cite{wang2021neus,yariv2021volume,oechsle2021unisurf}, mesh-based~\cite{zhang2021ners,goel2022differentiable}, or hybrid~\cite{chen2022tensorf,fridovich2022plenoxels} representations, these methods learn instance-specific representations capable of synthesizing novel views. However, as these methods do not learn generic data-driven priors, they typically require densely sampled views to be able to infer geometrically consistent underlying representations and are incapable of \emph{predicting} beyond what they directly observe.

\vspace{-2mm}
\paragraph{Projection-guided Generalizable View Synthesis.} Closer to our goal, several methods have aimed to learn models capable of view-synthesis across instances. While initial attempts~\cite{srn_sitzmann2019scene} used global-variable-conditioned neural fields, subsequent approaches~\cite{pixelnerf,mvsnerf,wang2021ibrnet,trevithick2021grf} obtained significant improvements by instead using features extracted via projection onto the context views. Reiznestein \etal~\cite{reizenstein2021common} further demonstrated the benefits of learning the aggregation mechanisms across the features along a query ray, but the projection-guided features remained the fundamental building blocks.
While these projection-based methods are effective at generating novel views by  transforming the visible structures, they  struggle to deal with large viewpoint changes (as the underlying geometry maybe uncertain), and are fundamentally unable to generate plausible visual information not directly observed in the context views. We argue that this is because these methods lack the mechanisms to  learn and utilize contexts globally when generating query views.

\begin{figure*}[!ht]
    \centering
    \includegraphics[width=1\linewidth]{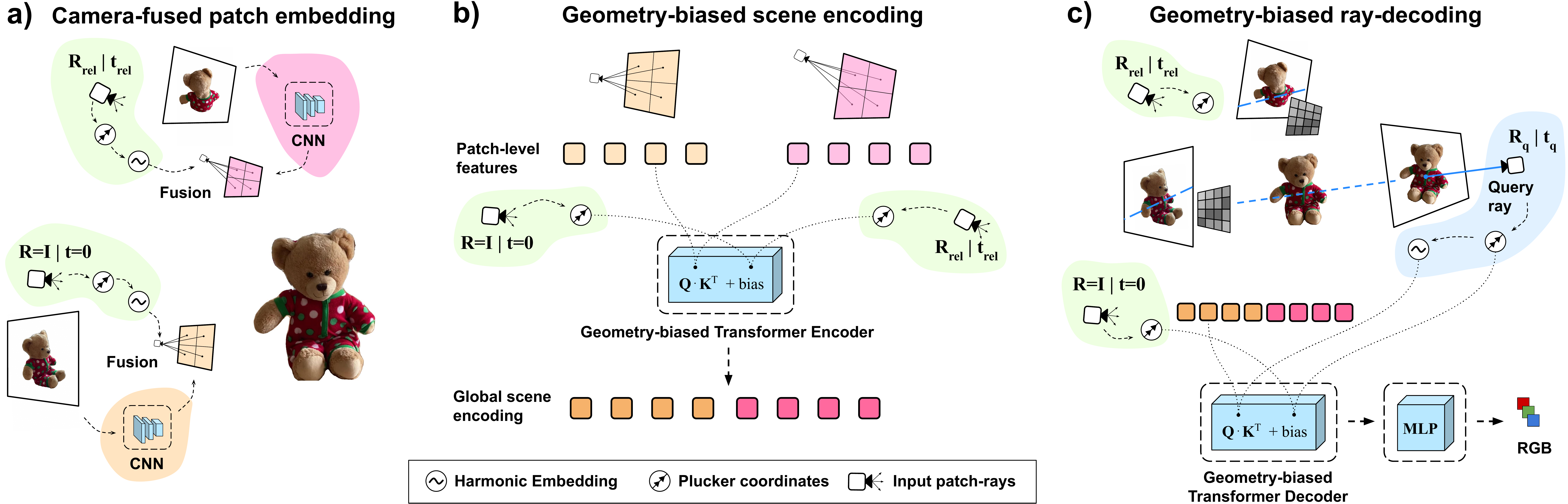}
    \caption{\small \textbf{Learning novel view synthesis using Geometry-biased Transformers.} Best viewed in color. \textbf{a) Camera-fused patch embedding}. Each input image $\Imat_i$ is processed using a shared CNN backbone $\fcnn$ and the feature maps are fused with the corresponding input patch-ray embeddings (obtained via $\pv_i$). \textbf{b) Geometry-biased scene encoding.} Our proposed Geometry-biased Transformer encoder $\fenc$ converts the set of patch-level feature tokens into a scene encoding via self-attention biased with ray distances. \textbf{c) Geometry-biased ray-decoding.} To decode pixels for a novel viewpoint, we construct ray queries that are decoded by a geometry-biased transformer decoder $\fdec$ by attending into the scene encoding. Finally, an MLP predicts the pixel color using the decoded query token.}
    \label{fig:approach}
\end{figure*}

\vspace{-2mm}
\paragraph{Geometry-free View Synthesis.} To allow using global context for view synthesis, an alternate class of methods uses `geometry-free' encodings to infer novel views. The initial learning-based methods~\cite{geometry_free_tatarchenko2015single,zhou2016view,geometry_free_yang2015weakly} typically focused on novel-view prediction given a single  image via global conditioning.
Subsequent approaches~\cite{gfvs,svs_transformer_gqn_nguyen2020sequential,viewformer} improved performance using different architectures \eg Transformers~\cite{vaswani2017attention}, while also allowing for probabilistic view synthesis using VQ-VAEs~\cite{vqvae_van2017neural} and VQ-GANs~\cite{vqgan_esser2021taming}. While this leads to detailed and realistic outputs, the renderings are not 3D-consistent due to stochastic sampling.

Our work is inspired by the recently proposed Scene Representation Transformer (SRT) \cite{srt}, which uses a set-latent representation that encodes both patch-level and global scene context. This design engenders a fast, deterministic rendering pipeline that, unlike projection-based methods, furnishes plausible hallucinations in the invisible regions. However, these benefits come at the cost of detail -- unlike the projection-based methods, this geometry-free approach is unable to capture precise details in the visible aspects. 
Motivated by this need to improve the detail,  we propose mechanisms to inject geometric biases in this framework, and find that this significantly improves the performance while preserving global reasoning and efficiency.

%%%%%%%%%%%%%%%%%%%%%%%%%%%

\vspace{-2mm}
\section{Approach}
\vspace{-2mm}
\label{approach}
We aim to render novel viewpoints of previously unseen objects from a few posed images. To achieve this goal, we design a rendering pipeline that reasons along the following two aspects: (i) \textbf{appearance} - \textit{what is the likely appearance of the object from the queried viewpoint}, and, (ii) \textbf{geometry} - \textit{what geometrically-informed context can be derived from the configuration of the given input and query cameras?}

Prior methods address each question in isolation \textit{e.g.} via global latent representations \cite{srt,wu2018learningshapepriors,viewformer,srn_sitzmann2019scene} that address (i) by learning object semantics, or, via reprojections \cite{pixelnerf,reizenstein2021common} that address (ii) by employing explicit geometric transformations. In contrast to prior works, our method jointly reasons along both these aspects. Concretely, we propose geometry-biased transformers that incorporate geometric inductive biases while learning set-latent representations that help capture global structures with superior quality.

Fig. \ref{fig:approach} depicts the Geometry-biased Transformer (GBT) framework which has three components. First, a shared CNN backbone extracts patch-level features which are fused with the corresponding ray embeddings to derive local (pose-aware) features (Fig. \ref{fig:approach}{\red{a}}). Then, the flattened patch features and the associated rays are fed as input tokens to the GBT encoder that constructs a global set-latent representation via self-attention (Fig. \ref{fig:approach}{\red{b}}). The attention layers are biased to prioritize both the photometric and the geometric context. Finally, the GBT decoder converts target ray queries to pixel colors by attending to the set-latent representation (Fig. \ref{fig:approach}{\red{c}}). We now review the preliminary concepts before describing our approach in detail.

\subsection{Preliminaries}
\label{prelim}

\begin{figure*}[!ht]
    \centering
    \includegraphics[width=0.95\linewidth]{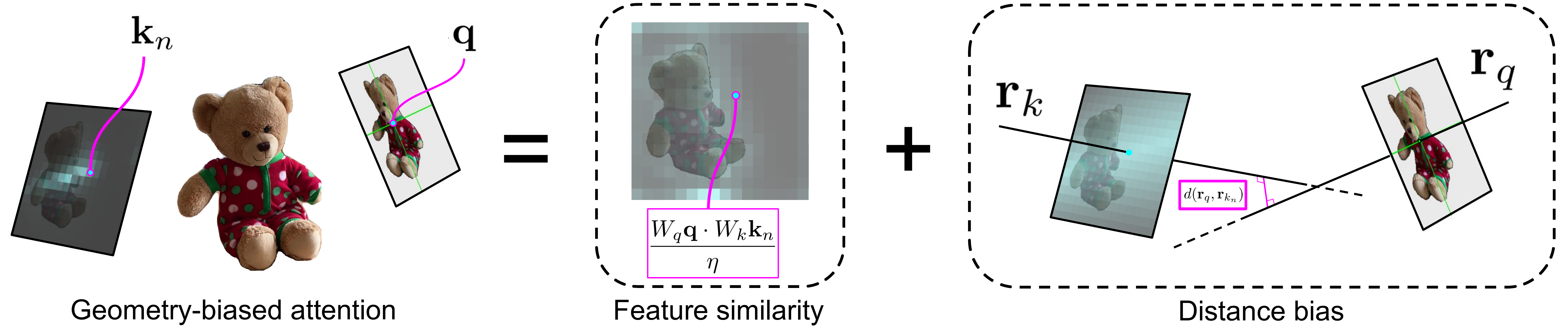}
    \caption{\textbf{An illustration of attention within GBT layer.} Given the query and key tokens $\qv$, $\kv_n$, along with the associated rays $\rv_q$, $\rv_{k_n}$, the attention within GBT incorporates two components: (i) a dot product similarity between features, and, (ii) the geometric distance bias computed between the rays. Refer to Eq. \ref{eq:attention_weights_gbt} for the exact computation. Best viewed in color.}
    \label{fig:epipolar_distance_bias}
\end{figure*}

\subsubsection{Ray representations}
The fundamental unit of geometric information in our approach is a ray which is used to compute the geometric similarity between two image regions.
A naive choice for ray representation is $\rv = (\ov, \dv)$, where $\ov \in \mathbb{R}^3$ is the origin of the ray, and $\dv \in \mathbb{S}^2$ is the normalized ray direction.

In contrast, we use the 4 DoF \plucker coordinates  \cite{lightfield_lfn_sitzmann2021light,jia2020plucker}, $\rv = (\dv, \mv) \in \mathbb{R}^6$, where $\mv = \ov \times \dv$, that are invariant to the choice of the origin along the ray. Intuitively, this allows us to associate a single color (pixel RGB) to the entire ray, agnostic to its origin. In practice, this simplification mitigates overfitting to the camera origin during training.

\subsubsection{Scene Representation Transformers}
\label{srt_prelim}
The overall framework of our approach is inspired by SRT \cite{srt} that proposes a transformer encoder-decoder network for novel view synthesis. Given a collection of posed images $\{(\Imat_i, \pv_i)\}_{i=1}^{V}$ where $\Imat \in \mathbb{R}^{H \times W \times 3}$
$\pv_i \in \mathbb{R}^{3\times4}$, and a query ray $\rv$, SRT computes the following:

\begin{equation}
\label{eq:srt_eq1}
    \{\zv_p\}_{p=1}^{V \times P} = \fenc \circ \fcnn(\{\Imat_i, \pv_i\})
\end{equation}
\vspace{-4mm}
\begin{equation}
    C(\rv) = \fdec( \rv \hspace{1mm} | \hspace{1mm} \{\zv_p\} )
\end{equation}

Here, the shared CNN backbone ($\fcnn$) extracts $P$ patch-level features from each posed input image. These are aggregated into a set of flat patch embeddings and fed as input tokens to the transformer encoder ($\fenc$). 
The encoder transforms input tokens into a set-latent scene representation $\{\zv_p\}$ via self-attention.
To render a novel viewpoint, the decoder $\fdec$ queries for each ray $\rv$ pertaining to the target pixels and yields an RGB color by attending to the scene representation $\{\zv_p\}$.

\subsection{Geometry-biased Transformer (GBT) Layer}

The core reasoning module in a transformer is a multi-head attention layer that aggregates information from the right context for each query. In our work, we propose to extend this module by incorporating geometric reasoning.

\noindent \textbf{Base transformer layer.} Given the query $\qv$, key $\{\kv_n\}$, value $\{\vv_n\}$ tokens, a typical transformer layer computes:

\vspace{-3mm}
\begin{align}
\label{eq:base_transformer_layer}
    \qv' = T(\qv, \{(\kv_n, \vv_n)\})
\end{align}
which consists of a multi-head attention module, followed by normalization and linear projection. During the context aggregating step, each multi-head attention layer aggregates token values based on query-key similarity weights:

\vspace{-3mm}
\begin{align}
\label{eq:attention_weights_classical}
    w_n & = \operatorname{softmax}_n \Big( \hspace{1mm} \frac{W_q \qv \cdot W_k \kv_n}{\eta} \hspace{1mm} \Big)
\end{align}

\paragraph{Incorporating ray distance as geometric bias.} In our use case, each query and context token pertains to some ray. For instance, all tokens passed to the encoder are patch embeddings that have associated patch rays (Fig. \ref{fig:approach}{\red{b}}). Likewise, we query the decoder using target pixel rays (Fig. \ref{fig:approach}{\red{c}}).

In such a scenario, we propose to bias the transformer's attention by encouraging similarity between rays that are closer to each other in 3D space. Specifically, the GBT layer couples the query and key tokens with the associated rays $(\qv, \rv_q), \{(\kv_n, \rv_{k_n})\}$ and performs the token transformation:

\vspace{-3mm}
\begin{align}
\label{eq:gbt_transformer_layer}
    \qv' = GBT((\qv, \rv_q), \{(\kv_n, \rv_{k_n}, \vv_n)\})
\end{align}

The attention layer is modified to account for the distance between $\rv_q = (\dv_q, \mv_q)$ and $\rv_{k_n} = (\dv_{k_n}, \mv_{k_n})$:

\vspace{-2mm}
\begin{align}
\label{eq:attention_weights_gbt}
    w_n & = \operatorname{softmax} \Big( \hspace{1mm} \frac{W_q \qv \cdot W_k \kv_n}{\eta} - \gamma^2 \hspace{1mm} d(\rv_q, \rv_{k_n}) \hspace{1mm} \Big)
\end{align}
\noindent where,
\begin{align}
\label{eq:ray_distance}
d (\rv_q, \rv_{k_n}) & = 
\begin{cases}
    \frac{|\dv_q \cdot \mv_{k_n} + \dv_{k_n} \cdot \mv_q|}{||\dv_q \times \dv_{k_n}||_2},& \dv_q \times \dv_{k_n} \ne 0 \\[2mm]
    \frac{||\dv_q \times (\mv_q - \mv_{k_n} / s)||}{||\dv_q||_2^2}, & \dv_{k_n} = s \dv_q, s \ne 0
\end{cases}
\end{align}

\noindent and $\gamma$ is a learnable parameter controlling the relative importance of geometric bias. This formulation explicitly accounts for both appearance (feature similarity between $\qv$ and $\kv_n$), and geometry (distance between $\rv_q$ and $\rv_{k_n}$). This attention mechanism is illustrated in Fig. \ref{fig:epipolar_distance_bias}.
In practice, the distance bias results in faster convergence to the right context during training. While one can fix $\biasweight$ to some constant hyperparameter, we found improved results by learning $\biasweight$.

\subsection{Learning Novel View Synthesis with GBTs}
\label{sec:arch_overview}

Given multiview images $\{\Imat_i \in \mathbb{R}^{H \times W \times 3}\}_{i=1}^V$ with paired camera poses $\{\pv_i \in \mathbb{R}^{3 \times 4}\}_{i=1}^V$, we wish to render a target viewpoint described by the camera pose $\pv_q \in \mathbb{R}^{3 \times 4}$. Our network, as illustrated in Fig. \ref{fig:approach}, first processes the posed multiview images using a CNN $\fcnn$ to extract patch-level latent features.
We then use GBT encoder $\fenc$ to extract a scene encoding, and GBT decoder $\fdec$ to yield pixel colors given target ray queries.

\begin{table*}[!ht]
    \centering
    \caption{\textbf{Evaluation of novel view synthesis.} Given $V=3$ input views, we evaluate the reconstruction quality (PSNR $\uparrow$ and LPIPS $\downarrow$) of each method on the  CO3Dv2\cite{reizenstein2021common} dataset. GBT denotes our proposed approach, and GBT-nb is an ablation. See Sec. \ref{sec:results}.
    }
    \resizebox{\linewidth}{!}{%
    \def\arraystretch{1.2}
    \setlength{\tabcolsep}{3pt}
    \begin{tabular}{ l  c c  c c  c c  c c  c c  c c  c c  c c  c c  c c  c c }
    \toprule
    \multirow{2}{*}{\textbf{10 training cat.}} & \multicolumn{2}{c}{\textbf{Apple}} & \multicolumn{2}{c}{\textbf{Ball}} & \multicolumn{2}{c}{\textbf{Bench}} & \multicolumn{2}{c}{\textbf{Cake}} & \multicolumn{2}{c}{\textbf{Donut}} & \multicolumn{2}{c}{\textbf{Hydrant}} & \multicolumn{2}{c}{\textbf{Plant}} & \multicolumn{2}{c}{\textbf{Suitcase}} & \multicolumn{2}{c}{\textbf{Teddybear}} & \multicolumn{2}{c}{\textbf{Vase}} & \multicolumn{2}{c}{\textbf{Mean}} \\
    \cmidrule(lr){2-3} \cmidrule(lr){4-5} \cmidrule(lr){6-7} \cmidrule(lr){8-9} \cmidrule(lr){10-11} \cmidrule(lr){12-13} \cmidrule(lr){14-15} \cmidrule(lr){16-17} \cmidrule(lr){18-19} \cmidrule(lr){20-21} \cmidrule(lr){22-23}
         & PSNR  & LPIPS  & PSNR  & LPIPS  & PSNR  & LPIPS  & PSNR  & LPIPS  & PSNR  & LPIPS  & PSNR  & LPIPS  & PSNR  & LPIPS  & PSNR  & LPIPS  & PSNR  & LPIPS  & PSNR  & LPIPS  & PSNR  & LPIPS  \\
         \midrule
         pixelNeRF \cite{pixelnerf} & 20.87 & 0.29 & 20.17 & 0.30 & 18.69 & 0.34 & 19.20 & 0.34 & 20.79 & 0.29 & 20.43 & 0.26 & 20.68 & 0.30 & 22.19 & 0.32 & 19.80 & 0.34 & 20.82 & 0.28 & 20.37 & 0.31 \\
         NerFormer \cite{reizenstein2021common}
         & 20.91 & 0.31 & 17.50 & 0.35 & 16.06 & 0.52 & 18.08 & 0.46 & 21.19 & 0.33 & 19.33 & 0.31 & 19.31 & 0.50 & 20.31 & 0.46 & 16.95 & 0.47 & 18.04 & 0.39 & 18.77 & 0.41 \\
         ViewFormer \cite{viewformer} & 21.70 & 0.24 & 19.34 & 0.30 & 17.08 & \textbf{0.30} & 18.04 & 0.32 & 19.59 & 0.28 & 18.59 & \textbf{0.21} & 18.34 & 0.31 & 21.61 & \textbf{0.26} & 16.60 & 0.31 & 21.52 & \textbf{0.21} & 19.24 & \textbf{0.27} \\
         \midrule
         GBT-nb & 22.83 & 0.28 & 20.59 & 0.32 & 19.22 & 0.34 & 20.56 & 0.34 & 21.87 & 0.31 & 21.32 & 0.24 & 21.52 & 0.30 & 23.30 & 0.29 & 19.82 & 0.34 & 22.65 & 0.27 & 21.37 & 0.30 \\
         GBT & \textbf{25.08} & \textbf{0.23} & \textbf{22.96} & \textbf{0.26} & \textbf{19.93} & 0.31 & \textbf{21.51} & \textbf{0.30} & \textbf{23.05} & \textbf{0.27} & \textbf{22.76} & 0.22 & \textbf{21.88} & \textbf{0.27} & \textbf{24.15} & 0.27 & \textbf{20.89} & \textbf{0.30} & \textbf{23.36} & 0.25 & \textbf{22.56} & \textbf{0.27}\\
         \bottomrule
    \end{tabular}
}
    \label{tab:10cat_expanded}
\end{table*}

\paragraph{a) Camera-fused patch embedding ($\fcnn)$.} We process each context image $\Imat_i$ through a ResNet18 backbone to obtain patch-level image feature grid. Subsequently, each patch feature is concatenated with the corresponding ray embedding (Fig. \ref{fig:approach}\red{a}) as follows:

\begin{equation}
\label{eq:gbt_eq_1}
    [\fv_c]_i^k = \Wm \Big( [\fcnn(\Imat_i)]^k  \oplus  h((\dv_i^k, \mv_i^k)) \Big)
\end{equation}

where $h(\cdot)$ denotes harmonic embedding \cite{mildenhall2020nerf}, $(\dv_i^k, \mv_i^k)$ denotes the \plucker coordinates for $k^{\text{th}}$ patch ray in the $i^{\text{th}}$ input image, and $\oplus$ denotes concatenation. We define each patch ray as the ray passing through the center of the receptive field of the corresponding cell in the feature grid. The concatenated features are projected using a linear layer $\Wm$.

While SRT fuses input images with per-pixel rays before the CNN, we fuse the CNN output feature grid with per-patch rays (observe different inputs to $\fcnn$ in Eq. \ref{eq:srt_eq1} and Eq. \ref{eq:gbt_eq_1}). This late fusion enables us to leverage transfer learning using pretrained image backbones. Furthermore, since the patch ray embeddings implicitly capture the positional information for each patch, we do not require 2D positional encoding or camera ID embedding after the CNN (unlike SRT), thus simplifying the architecture significantly.

\paragraph{b) Geometry-biased scene encoding ($\fenc$).} Given local patch features, we employ GBT encoder layers to augment them with the global scene context through self-attention. Specifically, we compute $\fv_e = \fenc(\fv_c, \{(\dv_i^k, \mv_i^k)\})$ where $\fenc$ contains a stack of GBT encoder layers as depicted in Fig. \ref{fig:approach}{\red{b}}. The query, key, and value tokens for the encoder layers are derived from the patch features $[\fv_c]_i^k$ and their corresponding patch rays $(\dv_i^k, \mv_i^k)$. For each transformer encoder layer, we learn a separate $\biasweight$ parameter.

Finally, the encoder outputs a global scene encoding $\{[\fv_e]_i^k\}$ that characterizes the appearance and the geometry of the object as observed from the multiple input views. Note, this extension of the set-latent representation \cite{srt} incorporates both appearance and geometric priors.

\paragraph{c) Geometry-biased ray decoding ($\fdec$).} To render a novel viewpoint given camera pose $\pv_q$, we construct an $H \times W$ grid of query rays $\rv_q=(\dv_q, \mv_q)$, with one ray per query pixel. We then employ a stack of GBT decoder layers $\fdec$ that decodes each query ray independently by aggregating meaningful context via cross-attention (Fig. \ref{fig:approach}{\red{c}}). Specifically, the query tokens for the multihead attention pertain to the query ray embeddings $h(\rv_q)$, while the keys and values comprise of the global scene encoding tokens $\{[\fv_e]_i^k\}$ along with the patch rays. The transformed query embeddings are processed by an MLP to predict the pixel color. Similar to $\fenc$, we learn a separate parameter $\biasweight$ for each GBT decoder layer in $F_D$.

\begin{figure*}[!t]
    \centering
    \includegraphics[width=1\linewidth]{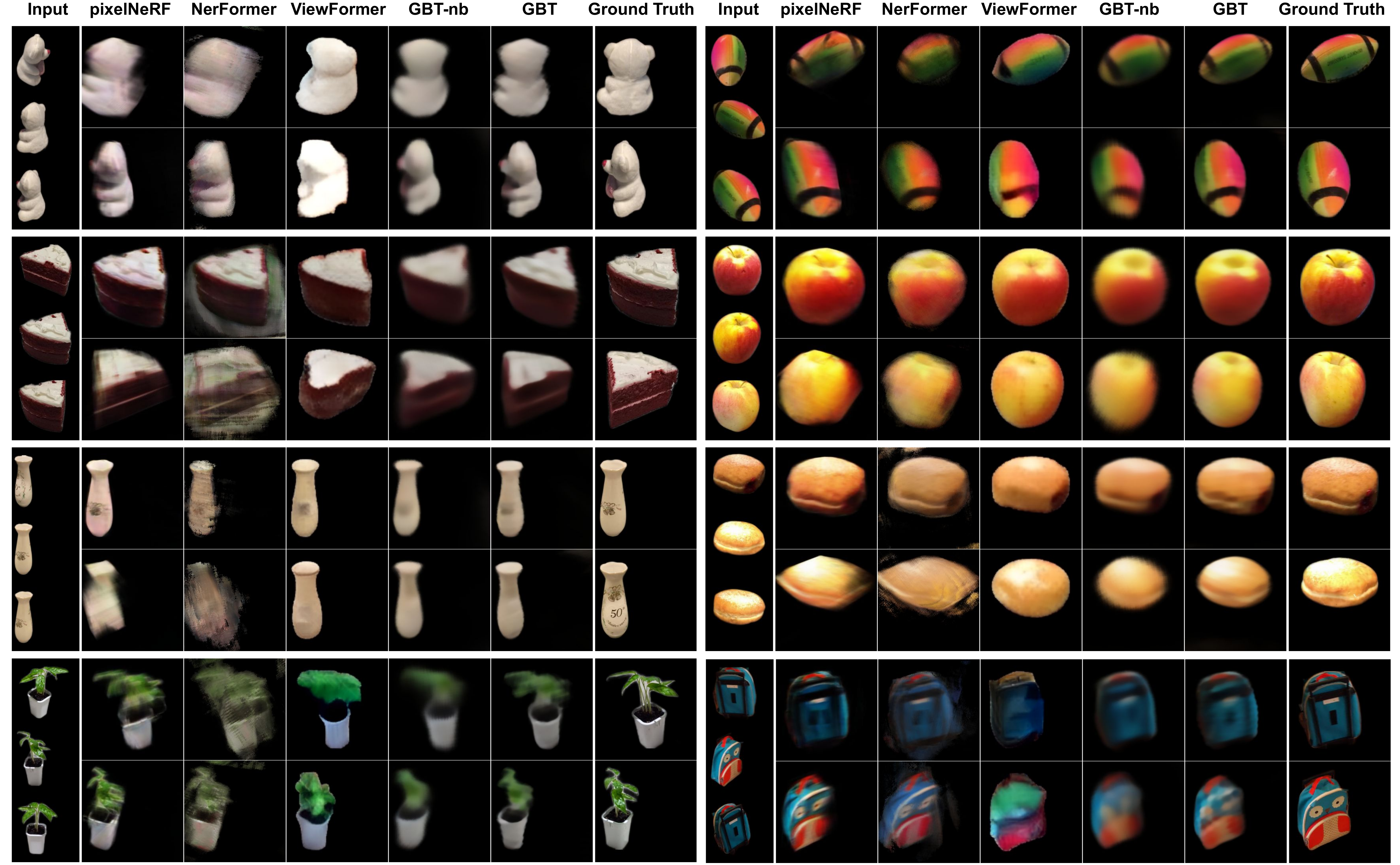}
    \caption{\textbf{Qualitative results on heldout objects from training categories.} For each object, we consider $V=3$ input views and compare the reconstruction quality of each method on 2 other query views. Best viewed in color.}
    \label{fig:main_figure}
\end{figure*}

\begin{table}[!t]
    \centering
    \caption {\textbf{Evaluation of variable context views setting.} We report PSNR ($\uparrow$) and LPIPS ($\downarrow$) averaged over 10 categories for each $V$.}
    \resizebox{\linewidth}{!}{%
    \begin{tabular}{ l c c c c c c }
    \toprule
    \multirow{2}{*}{\textbf{10 training cat.}} & \multicolumn{3}{c}{\textbf{PSNR $\uparrow$}} & \multicolumn{3}{c}{\textbf{LPIPS $\downarrow$}} \\
    \cmidrule(lr){2-4}  \cmidrule(lr){5-7}
         & $V=2$ & $V=3$ & $V=6$ & $V=2$ & $V=3$ & $V=6$ \\
         \midrule
         pixelNeRF \cite{pixelnerf} & 18.47 & 20.37 & 22.25 & 0.36 & 0.31 & \textbf{0.26} \\
         NerFormer \cite{reizenstein2021common} & 17.88 & 18.77 & 20.01 & 0.43 & 0.41 & 0.38 \\
         ViewFormer \cite{viewformer} & 18.62 & 19.24 & 20.12 & \textbf{0.28} & \textbf{0.27} & \textbf{0.26} \\
         \midrule
         GBT-nb & 20.91 & 21.37 & 21.49 & 0.31 & 0.30 & 0.30 \\
         GBT & \textbf{21.47} & \textbf{22.56} & \textbf{23.09} & 0.29 & \textbf{0.27} & 0.27 \\
         \bottomrule
    \end{tabular}
}
    \label{tab:10cat_avg}
\end{table}

\paragraph{Architectural details.} We use a ResNet18 (ImageNet initialized) up to the first 3 blocks as $F_C$. The images are resized to $H\times W = 256 \times 256$ and $F_C$ outputs a $16 \times 16$ feature grid. We use 8 GBT encoder layers and 4 GBT decoder layers, wherein each transformer contains 12 heads for multi-head attention with \textit{gelu} activation. For the harmonic embeddings $h$, we use $15$ frequencies $\{2^{-6}\pi, \ldots, 2^{8}\pi\}$. Since we do not have access to a consistent world coordinate frame across scenes, we choose an arbitrary input view as identity \cite{pixelnerf,srt}. 
All other cameras are represented relative to the identity view. See Appendix \ref{sec:architectural_details} for more details.

\paragraph{Training and Inference.} During training, we encode $V=3$ posed input views and query the decoder for $Q=7168$ randomly sampled rays for a given target pose $\pv_q$. The pixel color is supervised using an L2 reconstruction loss.  The model is trained with Adam optimizer with $10^{-5}$ learning rate until loss convergence. At inference, we encode the context views once and decode a batch of $H \times W$ rays for each query view in a single forward pass. This results in a fast rendering time. See Appendix \ref{sec:experimental_details} for more details.

%%%%%%%%%%%%%%%%%%%%%%%%%%%

\vspace{-2mm}
\section{Experiments}
\vspace{-2mm}
\label{experiments}
\subsection{Setup and Training Data}
\begin{table}[!t]
    \centering
    \caption{\textbf{Evaluation of novel-view synthesis on unseen categories}. Given $V=3$ input views, we evaluate the reconstruction quality (PSNR $\uparrow$ and LPIPS $\downarrow$) on unseen categories.}
    \resizebox{\linewidth}{!}{%
    \def\arraystretch{1.2}
    \setlength{\tabcolsep}{2pt}
    \begin{tabular}{ l  c c  c c  c c  c c  c c  c c}
    \toprule
    \multirow{2}{*}{\textbf{5 heldout cat.}} & \multicolumn{2}{c}{\textbf{Backpack}} & \multicolumn{2}{c}{\textbf{Book}} & \multicolumn{2}{c}{\textbf{Chair}} & \multicolumn{2}{c}{\textbf{Mouse}} & \multicolumn{2}{c}{\textbf{Remote}} & \multicolumn{2}{c}{\textbf{Mean}} \\
    \cmidrule(lr){2-3} \cmidrule(lr){4-5} \cmidrule(lr){6-7} \cmidrule(lr){8-9} \cmidrule(lr){10-11} \cmidrule(lr){12-13}
         & PSNR & LPIPS & PSNR & LPIPS & PSNR & LPIPS & PSNR & LPIPS & PSNR & LPIPS & PSNR & LPIPS \\
         \midrule
         pixelNeRF \cite{pixelnerf} & 22.87 & 0.31 & 18.86 & 0.34 & 20.30 & 0.32 & 23.39 & 0.27 & 23.74 & 0.23 & 21.83 & 0.30 \\
         ViewFormer \cite{viewformer} & 20.84 & 0.31 & 16.84 & \textbf{0.32} & 15.94 & 0.31 & 21.55 & 0.26 & 20.42 & 0.22 & 19.12 & 0.28 \\
         \midrule
         GBT-nb  & 23.55 & 0.33 & 19.38 & 0.35 & 20.50 & 0.32 & 23.72 & 0.27 & 24.00 & 0.22 & 22.23 & 0.30 \\
         GBT & \textbf{24.08} & \textbf{0.30} & \textbf{20.36} & \textbf{0.32} & \textbf{21.46} & \textbf{0.28} & \textbf{24.91} & \textbf{0.23} & \textbf{24.63} & \textbf{0.21} & \textbf{23.09} & \textbf{0.27}  \\
         \hline
    \end{tabular}
}
    \label{tab:5cat}
\end{table}

\paragraph{Dataset.} We experiment on the Common Objects in 3D (CO3Dv2) dataset \cite{reizenstein2021common} that contains multi-view images along with camera pose annotations. This is a challenging dataset containing real-world object captures from 51 MS-COCO categories. Following \cite{reizenstein2021common}, we train our network on 10 categories (see Table \ref{tab:10cat_expanded}). Further, we evaluate our method on 5 additional heldout categories (see Table \ref{tab:5cat}) to demonstrate generalization to unseen categories (see Appendix \ref{sec:experimental_details} for details on training and testing splits).

\paragraph{Baselines.} We benchmark GBT against three state-of-the-art methods:

\vspace{1mm}
\noindent - \textit{pixelNeRF} \cite{pixelnerf} which is a representative of projection-guided methods for generalizable view synthesis. 
Similar to our setting, we train a single category-agnostic pixelNeRF model on 10 categories from the CO3Dv2 dataset.

\vspace{1mm}
\noindent - \textit{NerFormer} \cite{reizenstein2021common} which uses attention-based mechanisms to aggregate projected features along a query ray. We utilize (category-specific) models provided by the authors. $^1$ \footnote{$^1$ While we evaluated  per-category models, the NerFormer authors conveyed this performance is similar to a  cross-category model.}

\vspace{1mm}
\noindent - \textit{ViewFormer} \cite{viewformer} which uses a two-stage `geometry-free' architecture to first encode the input images into a compact representation, and then uses a transformer model for view synthesis. For evaluation, we use the co3d-10cat model provided by the authors.

Additionally, we compare against another variant of our approach, where we replace the geometry-biased transformer layers with regular transformer layers (equivalently, set $\biasweight=0$ during training and inference). We refer to this as GBT-nb (no bias) in further discussion. GBT-nb is an extension of SRT \cite{srt}, where we use \plucker coordinates representation of rays and perform a late camera-fusion in the feature extractor.

\paragraph{Evaluation Metrics.} To evaluate reconstruction quality, we measure the peak signal-to-noise ratio (PSNR) and perceptual similarity metric (LPIPS). For each category, we select 10 scenes from the dev set for evaluation. We randomly sample $V$ context views and 32 query views for each scene and report the average metrics computed over these query views. We set appropriate seeds such that the context and query views are consistent across all methods.

\begin{table}[!t]
    \centering
    \caption{
    \textbf{Ablative analysis.} We train a separate category-specific model from scratch under each setting. The models are evaluated on the held out objects under consistent settings.}
    \resizebox{\linewidth}{!}{%
    \def\arraystretch{1.2}
    \setlength{\tabcolsep}{10pt}
    \begin{tabular}{ l  c c  c c }
    \toprule
    \multirow{2.5}{*}{\textbf{Method}} & \multicolumn{2}{c}{\textbf{Hydrant}} & \multicolumn{2}{c}{\textbf{Teddybear}} \\
    \cmidrule(lr){2-3} \cmidrule(lr){4-5}
         & PSNR ($\uparrow$) & LPIPS ($\downarrow$) & PSNR ($\uparrow$) & LPIPS ($\downarrow$) \\
         \midrule
         SRT*   & 19.63 & 0.23 & 19.48 & 0.32 \\
         GBT-nb  & 21.30 & 0.20 & 19.32 & 0.31 \\
         GBT-fb & 23.93 & \textbf{0.17} &  20.99 & 0.28 \\
         GBT & \textbf{24.22} & \textbf{0.17}  & \textbf{21.45} & \textbf{0.26} \\
         \bottomrule
    \end{tabular}
}
    \label{tab:ablation_ours}
\end{table}

\subsection{Results}
\label{sec:results}

\noindent \textbf{Novel view synthesis for unseen objects.} Table \ref{tab:10cat_expanded} demonstrates the efficacy of our method in synthesizing novel views for previously unseen objects. GBT consistently outperforms other methods in all categories in terms of PSNR. With the exception of a few categories, we also achieve superior LPIPS compared to other baselines.

For categories such as bench, hydrant, etc. we attribute ViewFormer's higher perceptual quality to their use of a 2D-only prediction model, which comes at the cost of multi-view consistent results. For instance, in Fig \ref{fig:main_figure}, ViewFormer's prediction for the donut is plausibly similar to some donut, however, lacks consistency with the corresponding ground truth query view. Also, in cases where the query view is not visible in any of the input views (ball, top-right), pixelNeRF and NerFormer - which rely solely on projection-based features from input images - suffer from poor results, while our method is capable of hallucinating these unseen regions.

Table \ref{tab:10cat_avg} analyses the performance of all methods with variable number of context views. While GBT is only trained with a fixed $V=3$ input views, it is capable of generalizing across different input view settings. We observe a higher performance gain under fewer context views (2-3). However, as the number of input views increases, pixelNeRF becomes more competitive.

\begin{figure}
    \centering
    \includegraphics[width=1\linewidth]{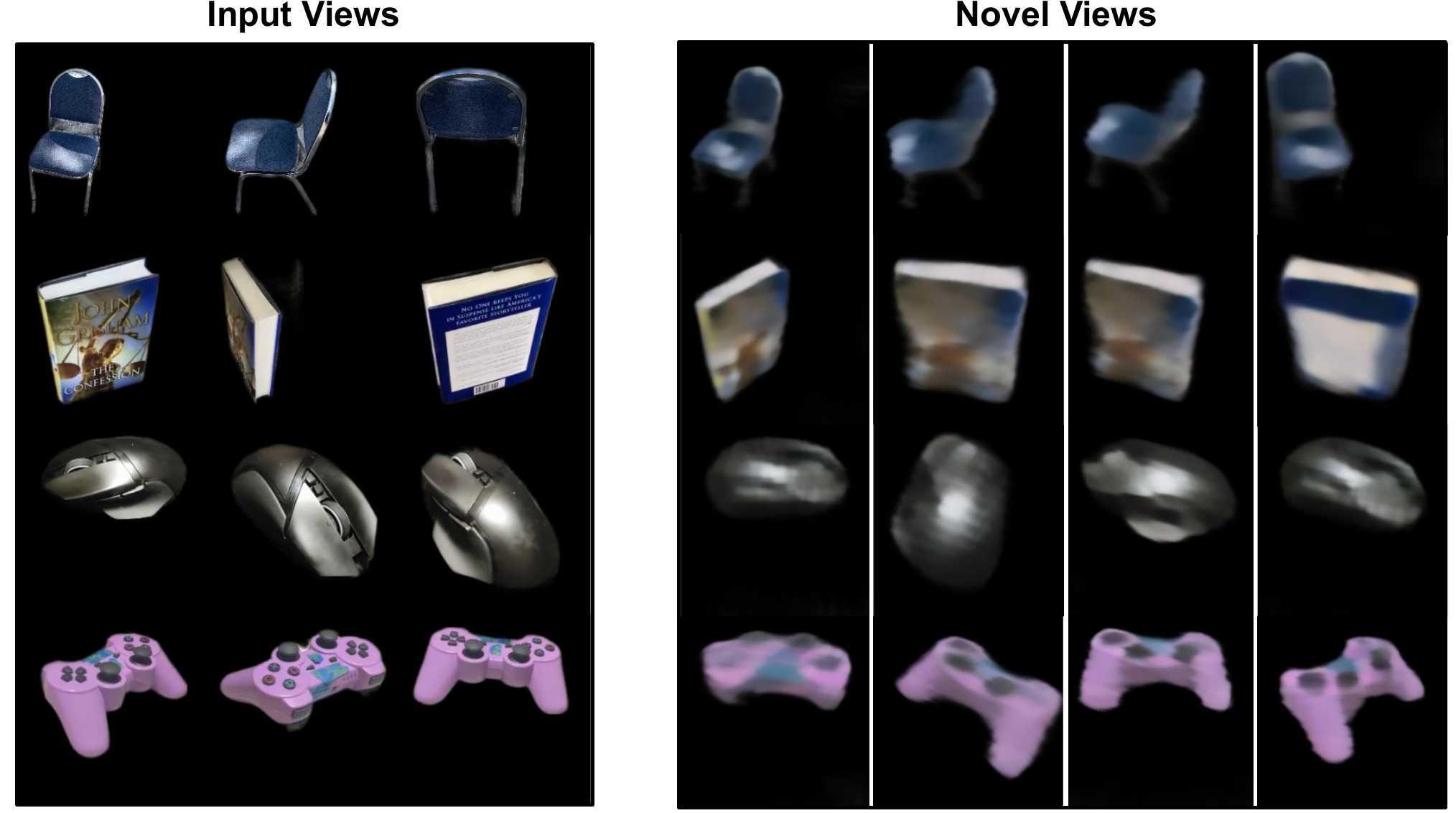}
    \caption{\textbf{Qualitative results on heldout categories.} On each row we visualize the rendered views obtained from GBT (right) given $V=3$ input views (left). Note that the model has never seen these categories of objects during training.}
    \label{fig:qualitative_5cat}
\end{figure}

\paragraph{Generalization to unseen categories.} To investigate whether our model learns generic 3D priors and can infer global context from given multi-view images, we test its ability to generalize to previously unseen categories. In Table \ref{tab:5cat} we benchmark our method by evaluating over 5 held out categories. We empirically find that GBT demonstrates better generalizability compared to baselines, and also observe this in the qualitative predictions in Figure \ref{fig:qualitative_5cat}.

\begin{figure}
    \centering
    \includegraphics[width=1\linewidth]{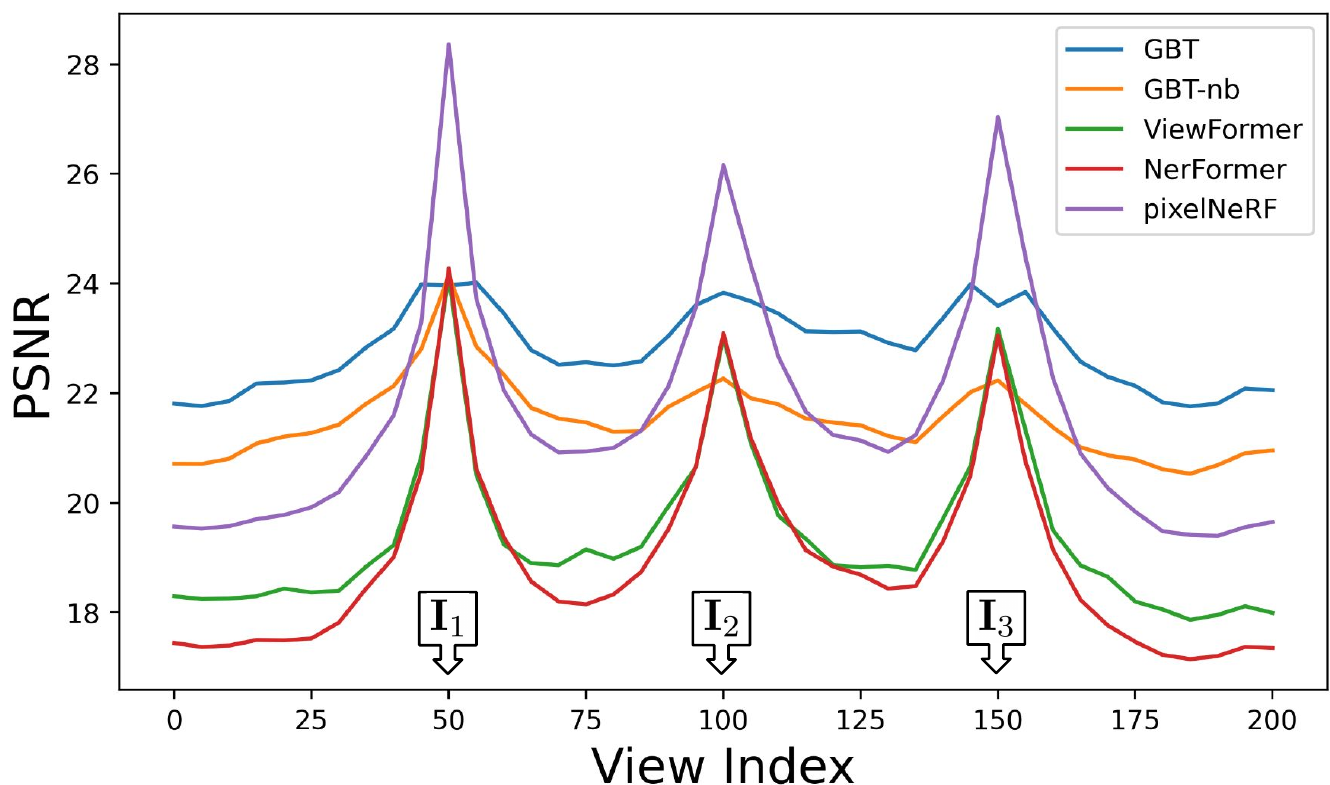}
    \caption{
    \textbf{Effect of viewpoint distance in prediction accuracy.} Given 200 frames, we set the $50^{th}, 100^{th}, 150^{th}$ frame as the input views, and evaluate the performance of novel view synthesis over all other views. While the prior methods show accurate results close to the input views, our approach (GBT) consistently outperforms them in other views.}
    \label{fig:viewpoint_analysis}
\end{figure}

\begin{figure*}[!ht]
    \centering
    \includegraphics[width=1\linewidth]{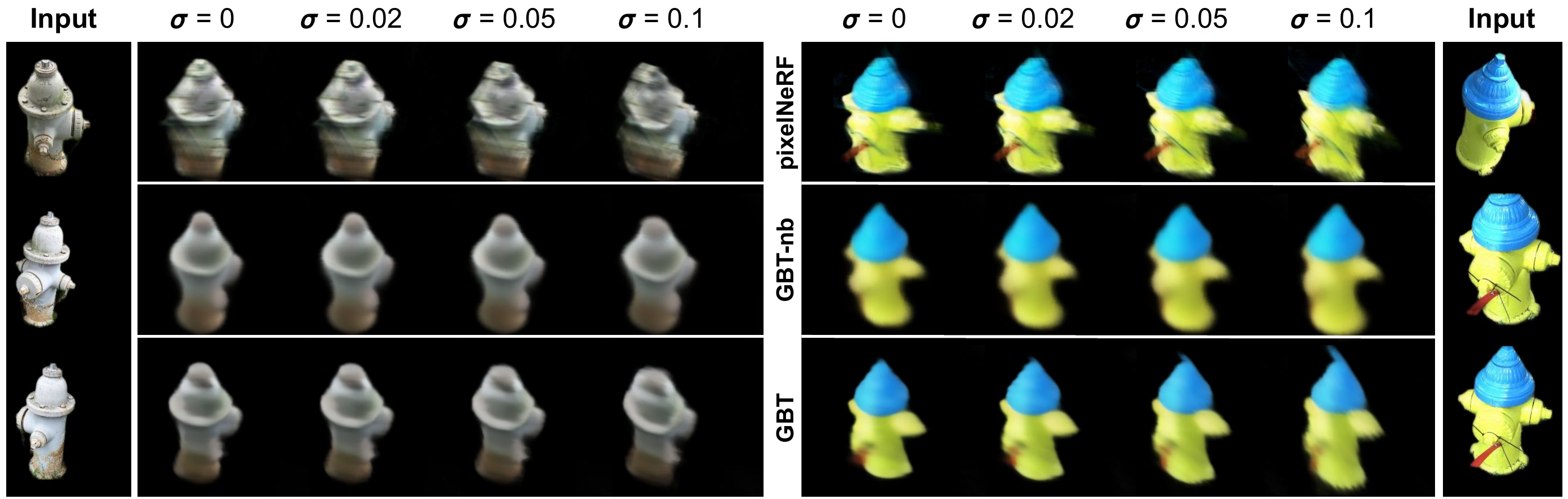}
    \caption{
    \textbf{Effect of camera noise.} 
    Given the 3 input views with noisy camera poses (increasing left to right), we visualize the predictions for a common query view across three methods (rows).}
    \label{fig:noisy_figure}
\end{figure*}

\subsection{Analysis}

\paragraph{Effect of Viewpoint Distance in Prediction Accuracy.}
In Fig \ref{fig:viewpoint_analysis}, we analyze view synthesis accuracy as a function of distance from context views. In particular, we use 80 randomly sampled sequences from across categories with 200 frames each, and set the $50^{th}, 100^{th}, 150^{th}$ views as context, and evaluate the average novel view synthesis accuracy across indices. We find that all approaches peak around the observed frames, but our set-latent representation based methods (GBT, GBT-nb) perform significantly better for query views dissimilar from the context views. This corroborates our intuition that a global set-latent representation is essential for reasoning in the sparse-view setup.

\begin{figure}
    \centering
    \includegraphics[width=1\linewidth]{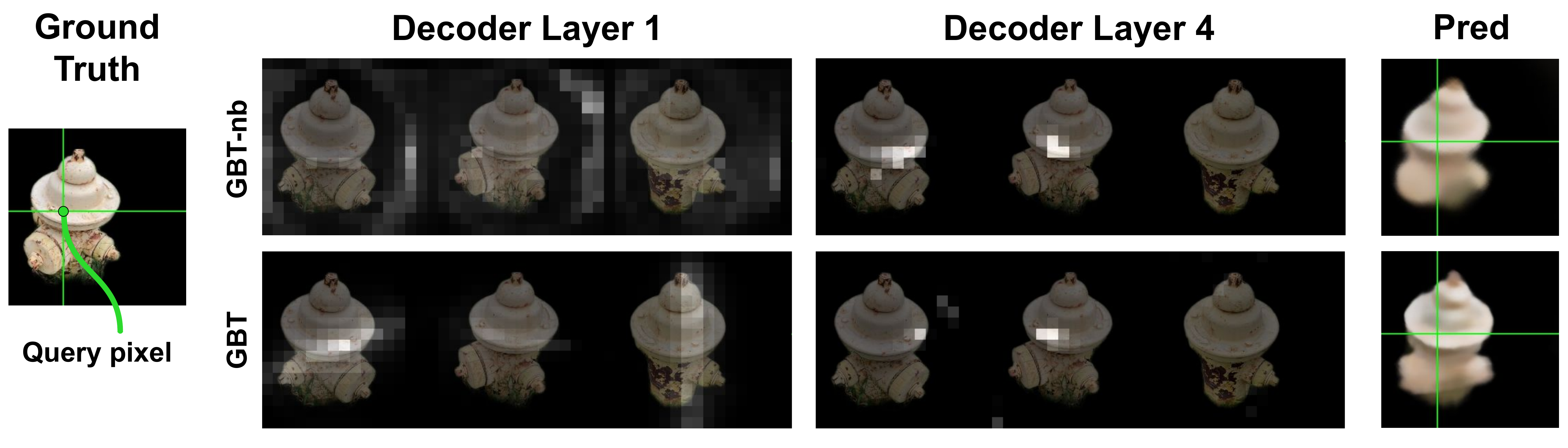}
    \caption{\textbf{Attention visualization.} For the query pixel marked in green, we visualize the attention over the input patches for the 1st and the 4th decoder layer. We compare the attention maps of GBT-nb (top) and GBT (bottom), wherein GBT is observed to yield sharper results. See Sec. \ref{sec:attention_vis}.}
    \label{fig:attention_figure}
\end{figure}

\paragraph{Ablative analysis.}
We investigate the importance of the design choices made in GBT, by ablating individual components and analysing performance.
First, we analyze the effect of learnable geometric bias by fixing $\gamma=1$ (GBT-fb) during the training process. Next, we remove the geometric bias component (GBT-nb); equivalently $\gamma=0$. Finally, we replace \plucker coordinates for ray representation with $\rv = (\ov, \dv)$. We term this trimmed version of GBT as SRT* (variant of SRT with late camera fusion).

For each ablation (see Table \ref{tab:ablation_ours}), we train a category-specific model from scratch and evaluate results on held-out objects. From Table \ref{tab:ablation_ours}, we see that learnable $\gamma$ yields some benefit over fixed $\gamma=1$. However, removing geometry altogether results in a considerable drop in performance. Also, the choice of \plucker coordinates as ray representations improves the predictions in general.

\paragraph{Robustness to camera noise.}
As the use of the geometric bias requires known camera calibration, we study the effect of noisy cameras on novel view synthesis. Following \cite{lin2021barf,srt}, we synthetically perturb input camera poses to various degrees and analyze the effect of noise during inference (for models trained without any camera noise during training).

We report the results in Table \ref{tab:noisy}, and see that performance degrades across all methods with camera noise. Although GBT-nb degrades more gracefully, the performance of GBT is better until a large amount of noise is added (about 10cm camera motion for a camera unit distance away from an object, and 9 degree rotation). Fig. \ref{fig:noisy_figure} demonstrates these observations visually.

\begin{table}[!t]
    \centering
    \caption{\textbf{Evaluation of noisy cameras.} All models are trained on 10 categories and evaluated on the Hydrant category.}
    \resizebox{\linewidth}{!}{%
    \def\arraystretch{1.4}
    \setlength{\tabcolsep}{2pt}
    \begin{tabular}{ l  c c  c c  c c  c c }
    \toprule
    & \multicolumn{2}{c}{$\sigma = 0$} & \multicolumn{2}{c}{$\sigma = 0.02$} & \multicolumn{2}{c}{$\sigma = 0.05$} & \multicolumn{2}{c}{$\sigma = 0.1$} \\
    \cmidrule(lr){2-3} \cmidrule(lr){4-5} \cmidrule(lr){6-7} \cmidrule(lr){8-9}
         & PSNR & LPIPS & PSNR & LPIPS & PSNR & LPIPS & PSNR & LPIPS\\
         \midrule
         pixelNeRF & 20.43 & 0.26 & 20.06 & 0.26 & 19.20 & 0.27 & 18.09 & 0.29 \\
         \midrule
         GBT-nb & 21.32 & 0.24 & 21.26 & 0.24 & 20.85 & 0.24 & \textbf{19.88} & \textbf{0.25} \\
         GBT & \textbf{22.76} & \textbf{0.22} & \textbf{22.40} & \textbf{0.22} & \textbf{21.43} & \textbf{0.23} & 19.84 & \textbf{0.25} \\
         \bottomrule
    \end{tabular}
}
    \label{tab:noisy}
\end{table}

\paragraph{Visualizing attention.}
\label{sec:attention_vis}

In Fig \ref{fig:attention_figure} we visualize attention heatmaps for a particular query ray highlighted in green.
In absence of geometric bias (GBT-nb), we observe a diffused attention map over the relevant context, which yields blurrier results. On adding geometric bias (GBT), we observe more concentrated attention toward the geometrically valid regions,  resulting in more accurate details.

%%%%%%%%%%%%%%%%%%%%%%%%%%%

\vspace{-2mm}
\section{Discussion}
\vspace{-2mm}
\label{discussion}
Our work introduced a simple but effective mechanism for adding geometric inductive biases in set-latent representation based networks. 
In particular, we demonstrated that for the task of novel view synthesis given few input views, this allows Transformer-based networks to better leverage geometric associations while preserving their ability to reason about global structure. While our approach led to substantial improvements over prior works, there are several unaddressed challenges. First, unlike projection-based methods, the set-latent representation methods (including ours) struggle to predict precise details and it remains on open question how one can augment such methods to overcome this. Moreover, the use of geometric information in our approach presumes access to (approximate) camera viewpoints for inference, and this may limit its applicability to in-the-wild settings. While our work focused on the task of view synthesis, we believe that the geometry-biasing mechanisms proposed would be relevant for other tasks where a moving camera is observing a common scene (\eg video segmentation, detection).

%%%%%%%%%%%%%%%%%%%%%%%%%%%
\paragraph{Acknowledgements.} We thank Zhizhuo Zhou, Jason Zhang, Yufei Ye, Ambareesh Revanur, Yehonathan Litman, and Anish Madan for helpful discussions and feedback. We also thank David Novotny and Jonáš Kulhánek for sharing outputs of their work and helpful correspondence. This project was supported in part by a Verisk AI Faculty Award.

%%%%%%%%% REFERENCES
{\small
\bibliographystyle{ieee_fullname}
\bibliography{references}
}

%%%%%%%%% APPENDIX
\newpage
\begin{appendices}

\section{Additional Random Results}
\label{sec:miscellaneous_results}

We provide additional results on randomly selected objects across each category, and, provide 360-degree rendering for each figure in the main text. See the project page for video visualizations, and Sec. \ref{suppl:sec:attention_visualization} for attention map visualizations on more examples across each category.

We observe that while ViewFormer produces plausible images, these are not 3d consistent due to the stochastic nature of the rendering pipeline.
While pixelNeRF and NerFormer produce accurate results in the vicinity of the observed context views, the results are inaccurate and implausible under larger camera deviations. 
Our baseline, GBT-nb produces consistent but blurry results. Finally, GBT improves over GBT-nb by furnishing finer details while preserving consistency across all viewpoints, although there is clear room for improvement in the level of details modeled.

\section{Classwise metrics for Table 2}

In Table \ref{tab:10cat_avg}, we present averaged results for $V=2, 3, 6$ over 10 categories. The per-category metrics are presented in Table \ref{tab:10cat_expanded_v2} (for $V=2$) and Table \ref{tab:10cat_expanded_v6} (for $V=6$). Note, the per-category results for $V=3$ setting is presented in the paper (in Table \ref{tab:10cat_expanded}).

\begin{table*}[!ht]
    \centering
    \caption{\textbf{Evaluation of novel view synthesis.} Given $V=2$ input views, we evaluate the reconstruction quality (PSNR $\uparrow$ and LPIPS $\downarrow$) of each method on the  CO3Dv2\cite{reizenstein2021common} dataset. GBT denotes our proposed approach, and GBT-nb is an ablation. 
    }
    \resizebox{\linewidth}{!}{%
    \def\arraystretch{1.2}
    \setlength{\tabcolsep}{3pt}
    \begin{tabular}{ l  c c  c c  c c  c c  c c  c c  c c  c c  c c  c c  c c }
    \toprule
    \multirow{2}{*}{\textbf{10 training cat.}} & \multicolumn{2}{c}{\textbf{Apple}} & \multicolumn{2}{c}{\textbf{Ball}} & \multicolumn{2}{c}{\textbf{Bench}} & \multicolumn{2}{c}{\textbf{Cake}} & \multicolumn{2}{c}{\textbf{Donut}} & \multicolumn{2}{c}{\textbf{Hydrant}} & \multicolumn{2}{c}{\textbf{Plant}} & \multicolumn{2}{c}{\textbf{Suitcase}} & \multicolumn{2}{c}{\textbf{Teddybear}} & \multicolumn{2}{c}{\textbf{Vase}} & \multicolumn{2}{c}{\textbf{Mean}} \\
    \cmidrule(lr){2-3} \cmidrule(lr){4-5} \cmidrule(lr){6-7} \cmidrule(lr){8-9} \cmidrule(lr){10-11} \cmidrule(lr){12-13} \cmidrule(lr){14-15} \cmidrule(lr){16-17} \cmidrule(lr){18-19} \cmidrule(lr){20-21} \cmidrule(lr){22-23}
         & PSNR  & LPIPS  & PSNR  & LPIPS  & PSNR  & LPIPS  & PSNR  & LPIPS  & PSNR  & LPIPS  & PSNR  & LPIPS  & PSNR  & LPIPS  & PSNR  & LPIPS  & PSNR  & LPIPS  & PSNR  & LPIPS  & PSNR  & LPIPS  \\
         \midrule
         pixelNeRF \cite{pixelnerf} & 18.21 & 0.36 & 17.74 & 0.35 & 17.59 & 0.38 & 17.22 & 0.38 & 18.51 & 0.35 & 18.44 & 0.31 & 19.39 & 0.36 & 20.71 & 0.37 & 17.74 & 0.41 & 19.17 & 0.34 & 18.47 & 0.36 \\
         NerFormer \cite{reizenstein2021common}
         & 20.11 & 0.34 & 16.63 & 0.37 & 15.09 & 0.55 & 17.23 & 0.48 & 20.07 & 0.36 & 18.11 & 0.35 & 18.37 & 0.53 & 19.69 & 0.46 & 15.73 & 0.51 & 17.79 & 0.39 & 17.88 & 0.43 \\
         ViewFormer \cite{viewformer} & 20.53 & 0.25 & 18.35 & 0.31 & 16.58 & 0.3 & 17.66 & 0.33 & 18.88 & \textbf{0.29} & 17.93 & \textbf{0.22} & 18.04 & 0.31 & 21.11 & \textbf{0.26} & 15.87 & \textbf{0.32} & 21.23 & \textbf{0.21} & 18.62 & 0.28 \\
         \midrule
         GBT-nb & 22.13 & 0.3 & 19.83 & 0.33 & 18.69 & 0.36 & 20.2 & 0.35 & 21.0 & 0.32 & 21.16 & 0.24 & 21.17 & 0.31 & 23.02 & 0.3 & 19.52 & 0.35 & \textbf{22.35} & 0.28 & 20.91 & 0.31 \\
         GBT & \textbf{22.96} & \textbf{0.27} & \textbf{21.45} & \textbf{0.28} & \textbf{19.1} & \textbf{0.33} & \textbf{20.71} & \textbf{0.32} & \textbf{21.78} & \textbf{0.29} & \textbf{21.82} & 0.23 & \textbf{21.29} & \textbf{0.29} & \textbf{23.41} & 0.28 & \textbf{19.93} & \textbf{0.32} & 22.28 & 0.26 & \textbf{21.47} & \textbf{0.29} \\
         \bottomrule
    \end{tabular}
}
    \label{tab:10cat_expanded_v2}
\end{table*}

\begin{table*}[!ht]
    \centering
    \caption{\textbf{Evaluation of novel view synthesis.} Given $V=6$ input views, we evaluate the reconstruction quality (PSNR $\uparrow$ and LPIPS $\downarrow$) of each method on the  CO3Dv2\cite{reizenstein2021common} dataset. GBT denotes our proposed approach, and GBT-nb is an ablation.
    }
    \resizebox{\linewidth}{!}{%
    \def\arraystretch{1.2}
    \setlength{\tabcolsep}{3pt}
    \begin{tabular}{ l  c c  c c  c c  c c  c c  c c  c c  c c  c c  c c  c c }
    \toprule
    \multirow{2}{*}{\textbf{10 training cat.}} & \multicolumn{2}{c}{\textbf{Apple}} & \multicolumn{2}{c}{\textbf{Ball}} & \multicolumn{2}{c}{\textbf{Bench}} & \multicolumn{2}{c}{\textbf{Cake}} & \multicolumn{2}{c}{\textbf{Donut}} & \multicolumn{2}{c}{\textbf{Hydrant}} & \multicolumn{2}{c}{\textbf{Plant}} & \multicolumn{2}{c}{\textbf{Suitcase}} & \multicolumn{2}{c}{\textbf{Teddybear}} & \multicolumn{2}{c}{\textbf{Vase}} & \multicolumn{2}{c}{\textbf{Mean}} \\
    \cmidrule(lr){2-3} \cmidrule(lr){4-5} \cmidrule(lr){6-7} \cmidrule(lr){8-9} \cmidrule(lr){10-11} \cmidrule(lr){12-13} \cmidrule(lr){14-15} \cmidrule(lr){16-17} \cmidrule(lr){18-19} \cmidrule(lr){20-21} \cmidrule(lr){22-23}
         & PSNR  & LPIPS  & PSNR  & LPIPS  & PSNR  & LPIPS  & PSNR  & LPIPS  & PSNR  & LPIPS  & PSNR  & LPIPS  & PSNR  & LPIPS  & PSNR  & LPIPS  & PSNR  & LPIPS  & PSNR  & LPIPS  & PSNR  & LPIPS  \\
         \midrule
         pixelNeRF \cite{pixelnerf} & 23.07 & 0.24 & 22.26 & \textbf{0.25} & 19.94 & \textbf{0.29} & 21.18 & \textbf{0.28} & 23.02 & \textbf{0.24} & 22.62 & 0.21 & 21.86 & \textbf{0.26} & 23.78 & 0.27 & \textbf{21.35} & 0.29 & 23.38 & 0.22 & 22.25 & \textbf{0.26} \\
         NerFormer \cite{reizenstein2021common}
         & 22.03 & 0.26 & 18.16 & 0.33 & 17.09 & 0.5 & 19.53 & 0.43 & 23.1 & 0.29 & 21.1 & 0.27 & 20.62 & 0.46 & 21.48 & 0.43 & 18.29 & 0.44 & 18.73 & 0.37 & 20.01 & 0.38 \\
         ViewFormer \cite{viewformer} & 22.66 & \textbf{0.23} & 20.11 & 0.29 & 18.06 & 0.28 & 19.05 & 0.31 & 20.79 & 0.27 & 19.62 & \textbf{0.2} & 18.94 & 0.29 & 22.18 & \textbf{0.25} & 17.57 & 0.29 & 22.2 & \textbf{0.21} & 20.12 & \textbf{0.26} \\
         \midrule
         GBT-nb & 22.53 & 0.28 & 20.59 & 0.32 & 19.5 & 0.34 & 20.77 & 0.34 & 22.15 & 0.3 & 21.24 & 0.23 & 21.83 & 0.3 & 23.43 & 0.29 & 19.85 & 0.34 & 23.0 & 0.26 & 21.49 & 0.30 \\
         GBT & \textbf{25.5} & \textbf{0.23} & \textbf{23.35} & 0.26 & \textbf{20.64} & 0.3 & \textbf{22.34} & 0.3 & \textbf{23.55} & 0.27 & \textbf{23.18} & 0.21 & \textbf{22.46} & 0.27 & \textbf{24.65} & 0.26 & 21.22 & 0.3 & \textbf{24.06} & 0.25 & \textbf{23.10} & \textbf{0.26} \\
         \bottomrule
    \end{tabular}
}
    \label{tab:10cat_expanded_v6}
\end{table*}

\section{Architectural Details}
\label{sec:architectural_details}

\textbf{We will make our implementation publicly available} for reproducibility. We also describe the implementation details of GBT here. Overall, GBT consists of 3 components - the CNN backbone $\fcnn$, GBT Encoder $\fenc$ and the GBT Decoder $\fdec$. The input to the model is a set of $V$ posed images $\{(\Imat_i, \pv_i)\}_{i=1}^V$, and $H \times W$ ray queries $\{\rv_j\}_{j=1}^{H\times W}$ generated using the target camera pose $\pv_q$. The model outputs RGB colors for each query ray, which are then reshaped to generate an image of size $H \times W \times 3$. 

We use PyTorch for model development. In the discussion below, tensor shapes are annotated in \texttt{monospace} font. We omit the batch dimension for simplicity. Across all models, the image size used is $H=W=256$. 

\subsection{GBT}

\paragraph{a) Camera-fused patch embedding ($\fcnn$).} We use a ResNet18 backbone (upto Res3 block) shared across input images to extract patch level features. Concretely, given the $V$ input images $\{\Imat_i\}_{i=1}^V$ of shape \texttt{(V, 3, 256, 256)}, the CNN outputs a feature grid \texttt{(V, 256, 16, 16)}. 

Each of the $16 \times 16$ cells in the feature grid corresponds to a receptive field in the input image. We associate each receptive field with a ray that passes through its center (called as `input patch ray' in the paper). Each input patch ray is represented in the \plucker coordinates $(\dv_i^k, \mv_i^k) \in \mathbb{R}^6$ - a tensor of shape \texttt{(V, 16, 16, 6)}, where the notation implies $i^{th}$ image's $k^{th}$ patch. We extract harmonic embeddings \cite{mildenhall2020nerf} over the \plucker coordinates, $h((\dv_i^k, \mv_i^k))$, using 15 frequencies $f=-6, \ldots 8$. Specifically, we get $h(x) = [sin(2^f \pi x), cos(2^f \pi x)]$ for each coordinate. This results in a $6 * 2 * 15 = 180$-d feature representation, yielding a ray embedding tensor of shape \texttt{(V, 16, 16, 180)}.

The CNN features $\{[\fcnn(\Imat_i)]^k\}$ and the ray embeddings $h((\dv_i^k, \mv_i^k))$ are concatenated along the channel dimension $\{[\fcnn(\Imat_i)]^k \oplus h((\dv_i^k, \mv_i^k))\}$ that results in a tensor of shape \texttt{(V, 16, 16, 436)}. Finally, these features are projected to a $768$ dimensional feature space using a linear layer $\Wm$ (\textit{i.e.} camera fusion). The output of the first stage is therefore camera-fused patch level features $[\fv_c]^k_i$ represented by a tensor of shape \texttt{(V, 16, 16, 768)}.

\paragraph{b) Geometry-biased scene encoding.} We use GBT encoder to embed the global scene context into the patch features. The GBT encoder consists of 8 geometry-biased transformer encoder layers with GELU activation, 12 MHA heads, and 768-d latent feature size. Each MHA module is biased using ray distances as done in Eq. \ref{eq:attention_weights_gbt}.

We construct the query, key and value tokens using flattened patch embeddings. Each query and key token is associated with the patch ray (\plucker coordinates). Therefore, the input to the GBT encoder is patch-level feature tensor of shape \texttt{(V * 16 * 16, 768)} along with the patch ray tensor of shape \texttt{(V * 16 * 16, 6)}. Note, the patch ray tensor is the same across all 8 GBT encoder layers, while the learnable weight $\gamma$ is different for each layer.

The output of the GBT encoder module $\{[\fv_e]^k_i\}$ is a tensor of shape \texttt{(V * 16 * 16, 768)} which is the set-latent representation of the scene. Each output token $[\fv_e]^k_i$ summarizes the appearance and the geometry of the scene incorporating both local and global features. These output tokens are used as the memory for the GBT decoder module to decode ray queries as described below.

\paragraph{c) Geometry-biased ray decoding.} To render an image, we construct $Q$ ray queries using the query camera pose $\pv_q$ and use the GBT decoder to predict the RGB color for each pixel. The GBT decoder contains a stack of 4 geometry-biased transformer decoder layers, followed by a shallow MLP. Similar to encoder, each decoder layer consists of 12 MHA heads biased with ray distances, 768-d latent dimensions and GELU activation. The MLP consists of 2 ReLU activated hidden layers (256-d, 64-d) and a sigmoid activated output (0-1 normalized RGB values).

The decoder's query tokens consist of harmonic ray embeddings $h((\dv_j, \mv_j))$ and the \plucker coordinates $(\dv_j, \mv_j)$ for each query ray. Similar to the encoder, we use 15 frequencies which results in a harmonic ray embedding tensor of shape \texttt{(Q, 180)}. These are projected to a 768-d feature space (GBT decoder's input dimension) via a linear layer. The keys and values tokens (\textit{i.e.} memory) pertain to the set-latent representation output by the GBT encoder, \textit{i.e.} a tensor of shape \texttt{(V * 16 * 16, 768)}.

The GBT decoder outputs a tensor of shape \texttt{(Q, 768)} that consists of decoded ray features for each target pixel. Finally, the MLP predicts a tensor of shape \texttt{(Q, 3)} containing the RGB colors for each queried pixel. During training, we compute L2 reconstruction loss on $Q=7168$ predicted pixel colors, and at inference we predict the colors for $Q=256 \times 256$ rays which is reshaped to yield the image tensor of shape \texttt{(3, 256, 256)}.

\subsection{Ablations}

We also propose 3 ablations of GBT in the paper:

\paragraph{GBT-fb (fixed bias).} This variant employs a fixed $\gamma=1$ weight in all the geometry-biased transformer layers as opposed to learning the weight $\gamma$. During training this model requires lesser memory overhead since the gradients for $\gamma$ are no longer computed. At inference, the compute overhead is similar to GBT.

\paragraph{GBT-nb (no bias).} In this variant, we remove the geometry-biased transformer layers in the encoder and decoder, and replace them with regular transformer layers (implemented in PyTorch). During training and inference, this model incurs lesser computational overhead than GBT since the ray distances are no longer computed. However, this comes at the cost of quality, which corroborates the need to account for geometry during attention.

\paragraph{SRT*.} This variant is closest to SRT, wherein we no longer use geometric bias, nor \plucker ray representations. Rays are represented using the origin $\ov$ and direction $\dv$ as $\rv = (\ov, \dv)$. While the compute overhead is similar to GBT-nb, this model is the least performing among all the variants which demonstrates the benefits of our design choices.

\section{Experimental Details}
\label{sec:experimental_details}

\subsection{Training \& Inference}

\paragraph{Training.} We perform mixed-precision training with $2\times$NVIDIA A6000 (48GB) GPUs with a batch size of $B=6$ scenes. For each scene in a batch, we randomly sample $V=3$ input views and $Q=7168$ rays from an arbitrary query viewpoint. The predicted pixel RGB color for each query ray is supervised using an L2 reconstruction loss with respect to the ground truth pixel in the query viewpoint. The training is performed till loss convergence which is about 1.6Mil iterations for GBT and about 2Mil iterations for GBT-nb trained on all 10 categories (about 9-10 days).

\paragraph{Inference.} At inference we are provided with $V$ posed input images and a query camera pose $\pv_q$. We generate a batch of $H \times W = 256 \times 256$ query rays that are decoded in a single forward pass. The inference time for a single query image with $V=3$ input views for GBT is 0.09s ($\sim$ 11 FPS), and for GBT-nb is 0.025s ($\sim$ 40 FPS). Compared to GBT, the prior methods exhibit more runtime - pixelNeRF takes 7.3s ($\sim$ 0.13 FPS), NerFormer takes 2.7s ($\sim$ 0.37 FPS), and ViewFormer takes 0.68s ($\sim$ 1.5 FPS), using default parameters ($1 \times$A6000 GPU).

\subsection{Dataset Splits}

We use the CO3Dv2 dataset \cite{reizenstein2021common} that contains multiview images along with camera pose annotations of 51 object categories. We select 10 categories to train our models - \texttt{[apple, ball, bench, cake, donut, hydrant, plant, suitcase, teddybear, vase]}. Additionally, we choose 5 heldout categories - \texttt{[backpack, book, chair, mouse, remote]}, which are used to evaluate the generalization of methods. All images are cropped and resized to $256 \times 256$ (the camera parameters are modified accordingly).

CO3Dv2 provides three dataset splits - \texttt{fewview\_train}, \texttt{fewview\_dev}, and \texttt{fewview\_test}. Since the \texttt{fewview\_test} ground-truth has been redacted for online evaluation, we use \texttt{fewview\_train} for training and \texttt{fewview\_dev} for testing. We use all available views in each scene in \texttt{fewview\_train} split for training.

For computing metrics on the \texttt{fewview\_dev} split, we evaluate the models on 32 randomly selected views for the first 10 scenes in each category. We set random seed such that the input and query viewpoints are consistent across all methods. For the viewpoint distance experiment in Fig. \ref{fig:viewpoint_analysis}, we evaluate the average PSNR over 80 sequences across categories for each of 200 query views, with the $50^{\text{th}}, 100^{\text{th}}, 150^{\text{th}}$ views being the input views.

\section{Attention Visualization}
\label{suppl:sec:attention_visualization}

\begin{figure*}
    \centering
    \includegraphics[width=0.63\linewidth]{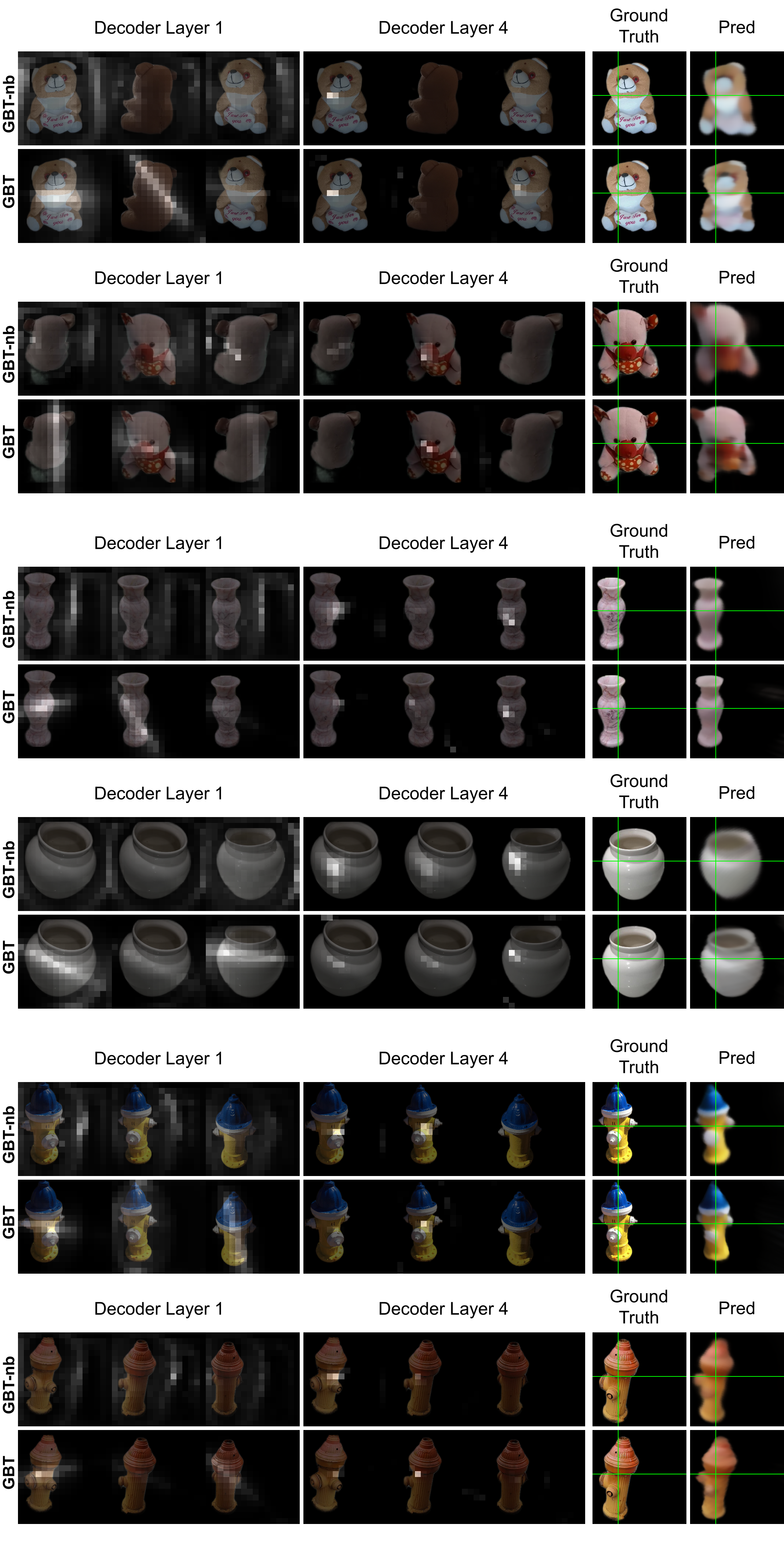}
    \caption{Attention maps for held out objects in teddybear, vase and hydrant categories.}
    \label{fig:attention_1}
\end{figure*}

\begin{figure*}
    \centering
    \includegraphics[width=0.63\linewidth]{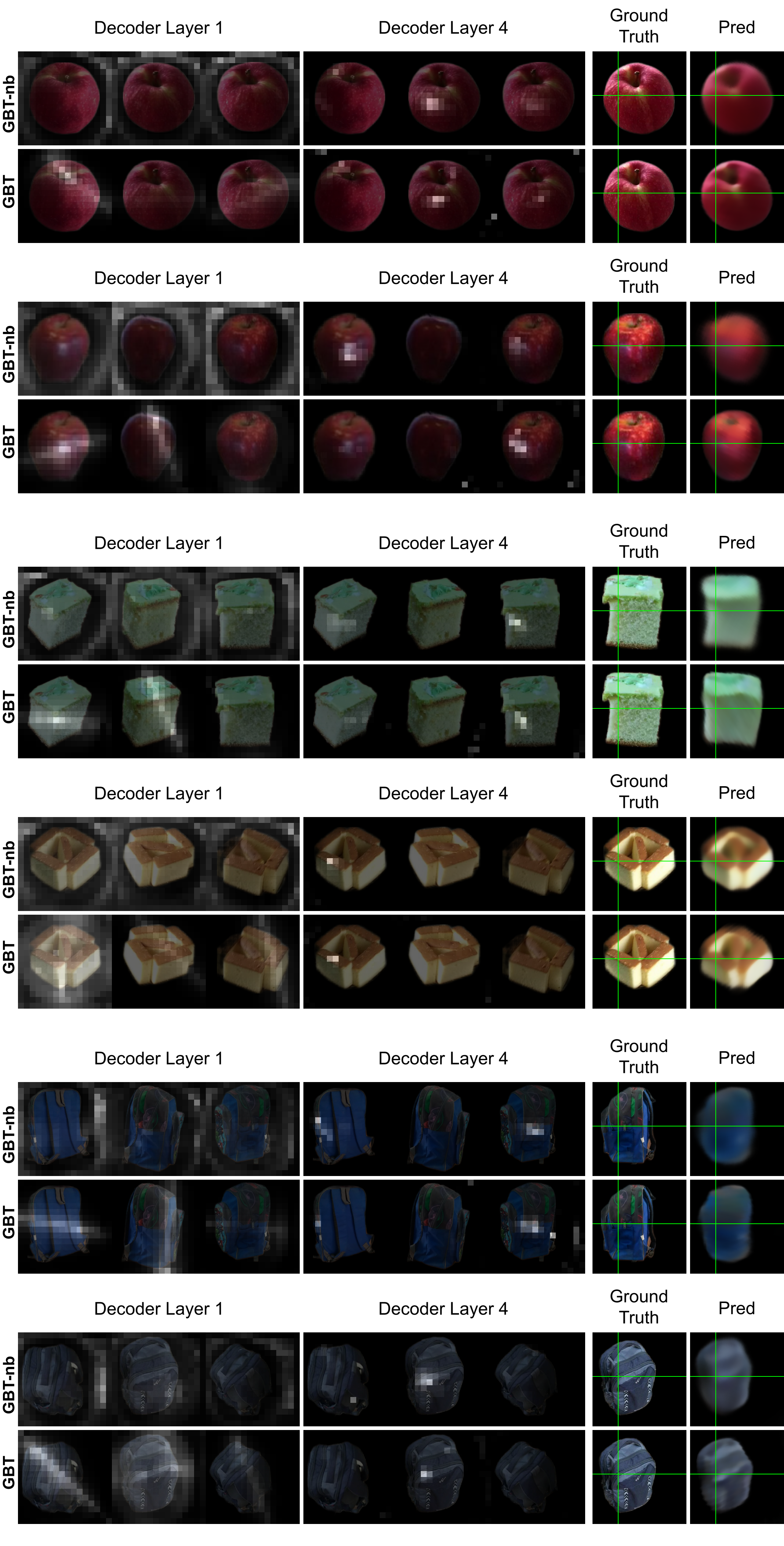}
    \caption{Attention maps for held out objects in apple, cake and backpack categories.}
    \label{fig:attention_2}
\end{figure*}

\begin{figure*}
    \centering
    \includegraphics[width=0.63\linewidth]{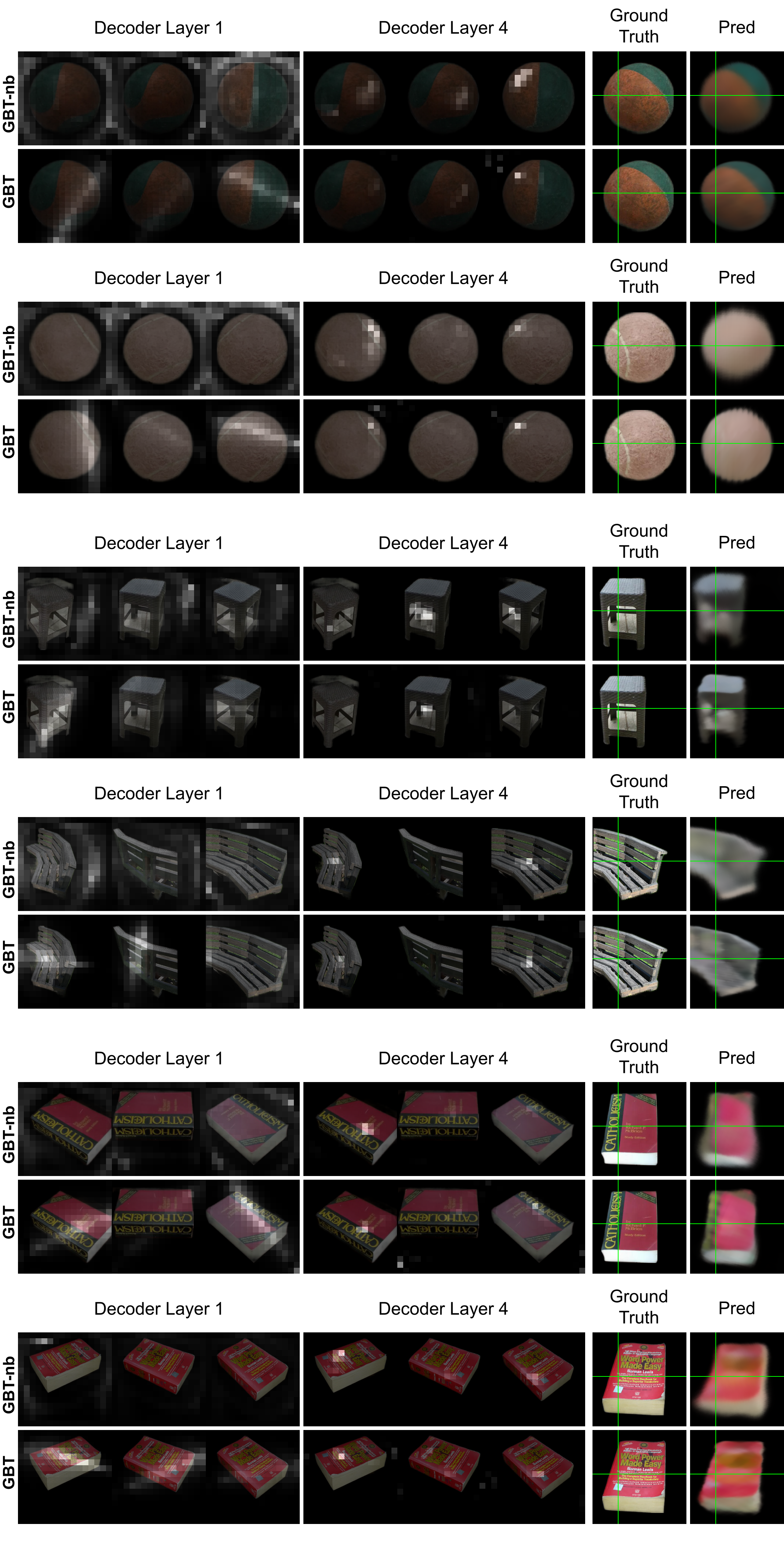}
    \caption{Attention maps for held out objects in ball, bench and book categories.}
    \label{fig:attention_3}
\end{figure*}

\begin{figure*}
    \centering
    \includegraphics[width=0.63\linewidth]{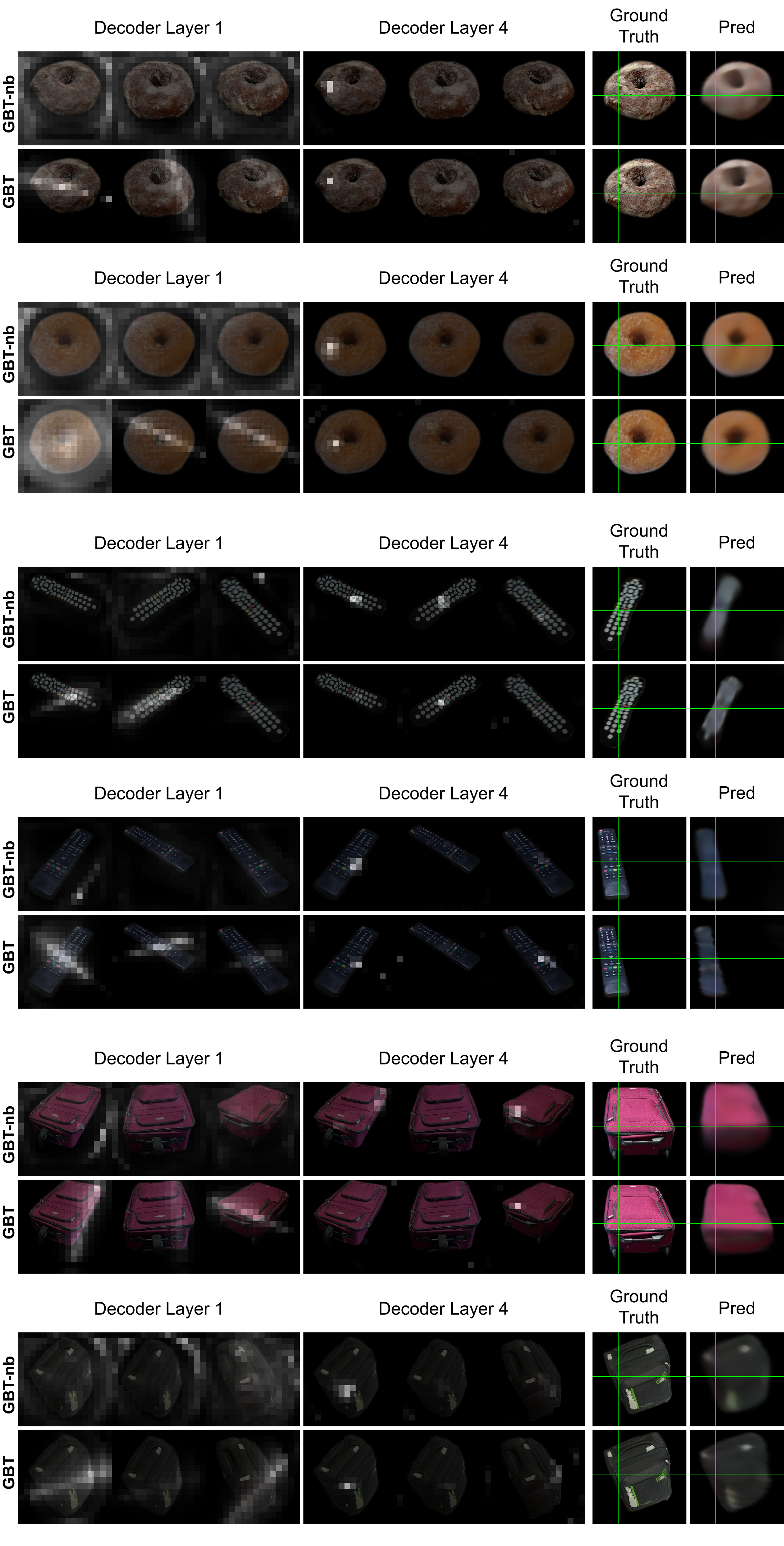}
    \caption{Attention maps for held out objects in donut, remote and suitcase categories.}
    \label{fig:attention_4}
\end{figure*}

\begin{figure*}
    \centering
    \includegraphics[width=0.63\linewidth]{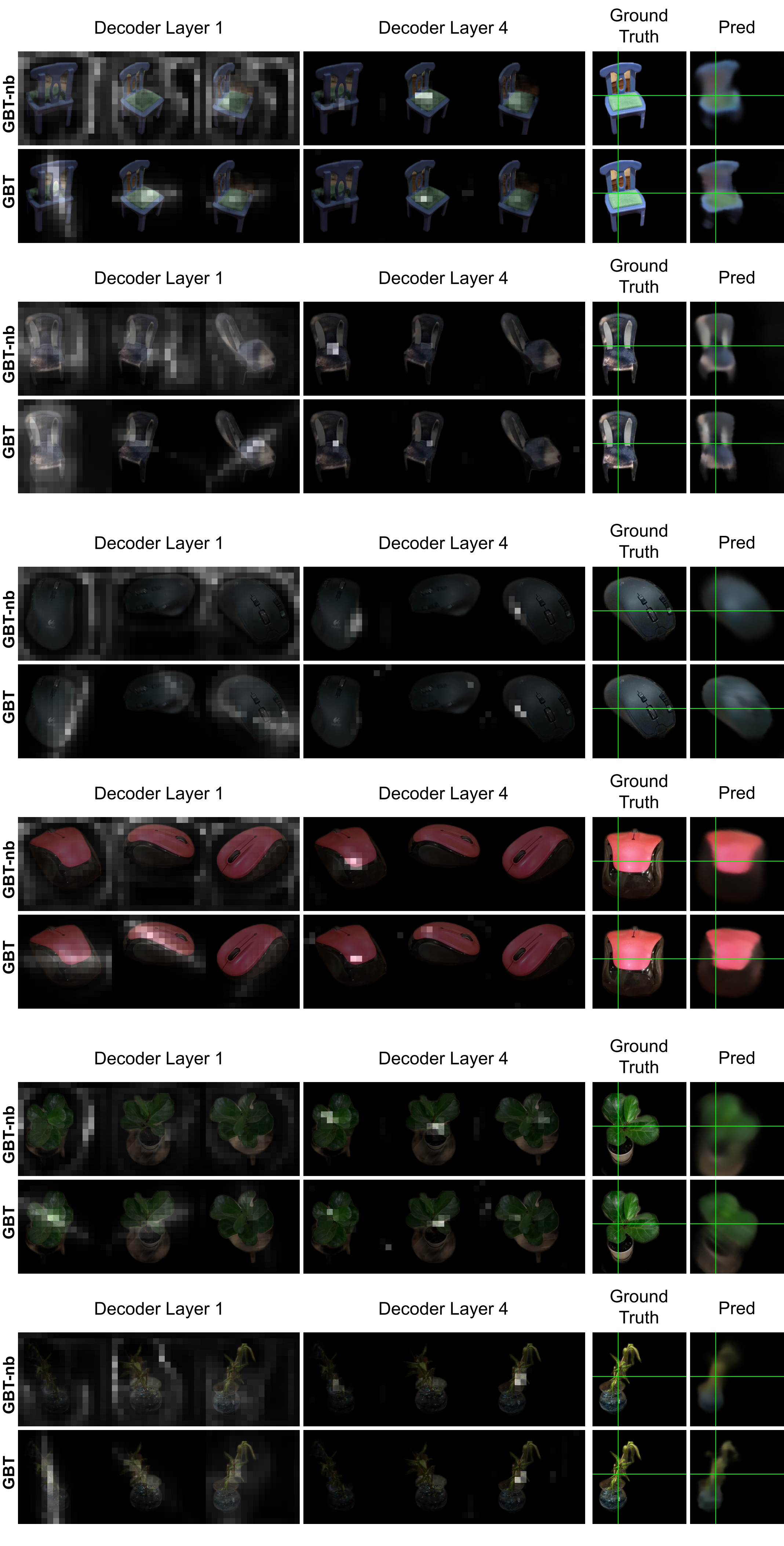}
    \caption{Attention maps for held out objects in chair, mouse and plant categories.}
    \label{fig:attention_5}
\end{figure*}

We plot the attention maps for GBT and GBT-nb in Fig. \ref{fig:attention_1}-\ref{fig:attention_5}. Overall, the incorporation of geometric bias results in more concentrated attention towards the geometrically valid regions. For instance, see the attention maps for GBT and GBT-nb in the two hydrant examples in Fig. \ref{fig:attention_1}. We hypothesize that concentrated attention toward the relevant context improves the quality of the rendered images.

% \newpage
% \newpage
% {\small
% \bibliographystyle{ieee_fullname}
% \bibliography{references}
% }

\end{appendices}

\end{document}